\documentclass[10pt,journal,compsoc]{IEEEtran}
\usepackage{cite}
\usepackage[pdftex]{graphicx}
\usepackage{amsmath}
\usepackage{array}
\usepackage{multirow}
\usepackage{xcolor}
\usepackage{bbm}
\usepackage{amsmath}
\usepackage{amssymb}
\usepackage{enumitem}
\usepackage{subcaption}
\usepackage{bbding}
\usepackage{pifont}
\newcommand{\xmark}{\ding{55}}
\definecolor{Olive_Green}{rgb}{0.0, 0.55, 0.0}

\usepackage[pagebackref=false,breaklinks=true,bookmarks=false]{hyperref}

\newcolumntype{P}[1]{>{\centering\arraybackslash}p{#1}}

\newcommand{\REVISION}[1]{\textcolor[rgb]{0.0,0.0,0.0}{#1}}
\newcommand{\SECONDRREVISION}[1]{\textcolor[rgb]{0.0,0.0,0.0}{#1}}

\begin{document}
\title{Multi-Scale 2D Temporal \REVISION{Adjacency} Networks for Moment Localization with Natural Language}
% \author{Songyang~Zhang,~Houwen~Peng,~Jianlong~Fu~and~Jiebo~Luo\IEEEcompsocitemizethanks{\IEEEcompsocthanksitem S. Zhang and J. Luo were with the Department
% of Computer Science, University of Rochester, Rochester,
% NY, 14627.\protect\\
% E-mail: szhang83@ur.rochester.edu
% \IEEEcompsocthanksitem H. Peng are J. Fu were with Microsoft Research.}
% }
\author{Songyang~Zhang,~Houwen~Peng,~Jianlong~Fu,~Yijuan Lu,~and~Jiebo~Luo\IEEEcompsocitemizethanks{\IEEEcompsocthanksitem S. Zhang and J. Luo are with the Department
of Computer Science, University of Rochester, Rochester,
NY, 14627.\protect\\
E-mail: {szhang83, jluo}@cs.rochester.edu
\IEEEcompsocthanksitem H. Peng, J. Fu and Y. Lu are with Microsoft. \protect\\
E-mail: \{houwen.peng,jianf, yijlu\}@microsoft.com
\IEEEcompsocthanksitem{Houwen~Peng is the corresponding author.}
}
}

\markboth{Journal of \LaTeX\ Class Files,~Vol.~14, No.~8, August~2015}%
{Shell \MakeLowercase{\textit{et al.}}: Bare Demo of IEEEtran.cls for Computer Society Journals}

\IEEEtitleabstractindextext{%
\begin{abstract}
%We address the problem of retrieving a specific moment from an untrimmed video by a query sentence. 
We address the problem of retrieving a specific moment from an untrimmed video by natural language. 
It is a challenging problem because a target moment may take place in the context of other temporal moments in the untrimmed video. 
% \HW{Existing methods cannot tackle this challenge well since they consider temporal moments individually and neglect the temporal dependencies. }
Existing methods cannot tackle this challenge well since they do not fully consider the temporal contexts between temporal moments.
In this paper, we model the temporal context between video moments by a set of predefined %\HW{several} 
two-dimensional maps under different temporal scales. For each map, one dimension indicates the starting time of a moment and the other indicates the duration. 
These 2D temporal maps can cover diverse video moments with different lengths, while representing their adjacent contexts at different temporal scales. 
Based on the 2D temporal maps, we propose a Multi-Scale Temporal \REVISION{Adjacency} Network (MS-2D-TAN), a single-shot framework for moment localization. 
It is capable of encoding the adjacent temporal contexts at each scale, while learning discriminative features for matching video moments with referring expressions. 
% A gated convolution is also employed to handle the non-rectangular shape of the 2D temporal map. 
We evaluate the proposed MS-2D-TAN on three challenging benchmarks, \textit{i.e.}, Charades-STA, ActivityNet Captions, and TACoS, where our MS-2D-TAN outperforms the state of the art.
\end{abstract}

%\begin{IEEEkeywords}
%Computer Society, IEEE, IEEEtran, journal, \LaTeX, paper, template.
%\end{IEEEkeywords}
}

\maketitle

\IEEEdisplaynontitleabstractindextext

\ifCLASSOPTIONpeerreview
\begin{center} \bfseries EDICS Category: 3-BBND \end{center}
\fi

\IEEEpeerreviewmaketitle

\IEEEraisesectionheading{\section{Introduction}\label{sec:introduction}}

\begin{figure*}
    \centering
    \includegraphics[width=\textwidth]{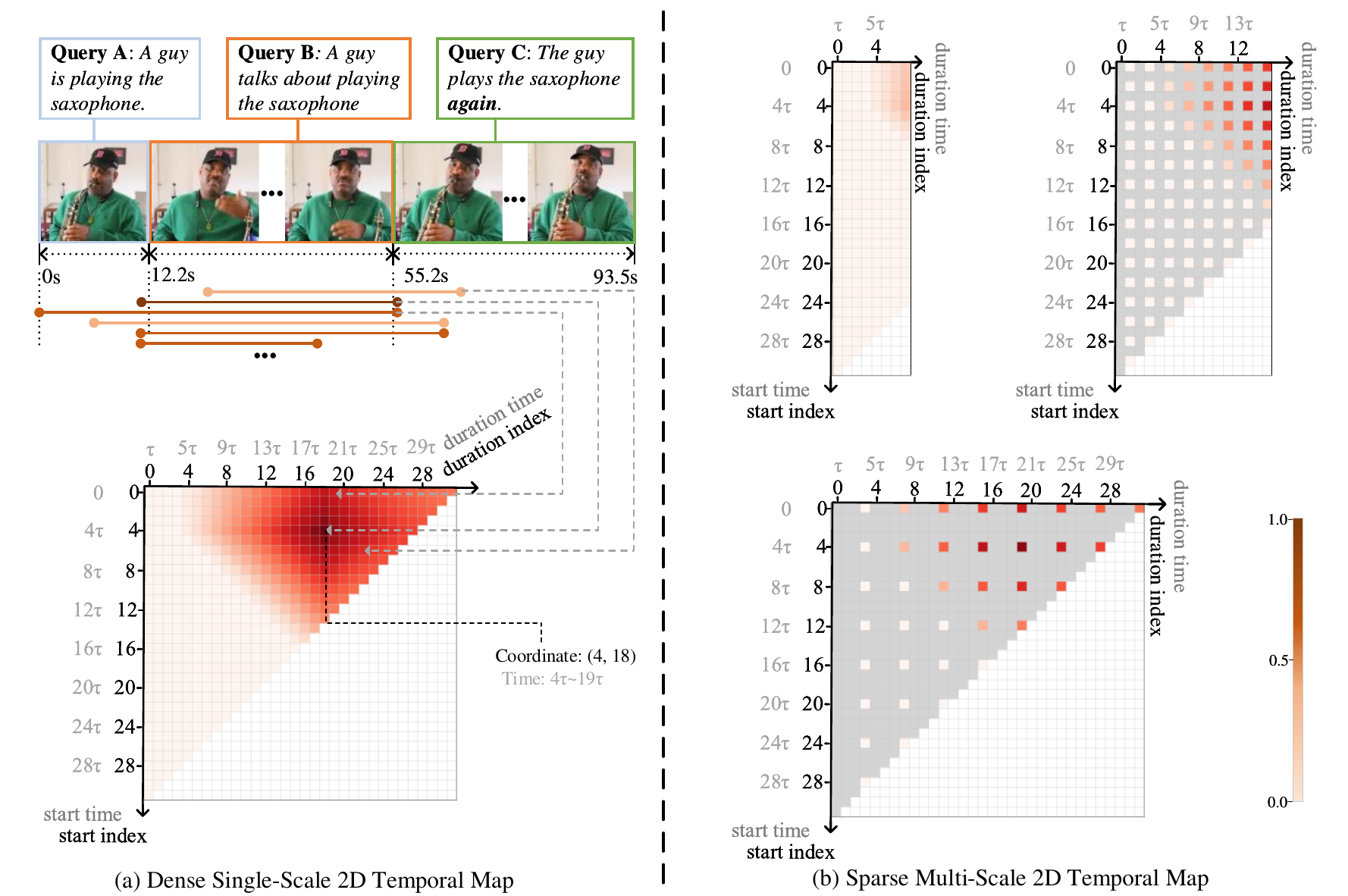}
    \caption{
%Examples of localizing moments with natural language in an untrimmed video. 
Examples of the 2D temporal map. 
(a) Dense Single-scale 2D Temporal Map: 
% \HW{(a) Single-scale Dense Map: } . (b) \HW{Multi-scale Sparse Map:.}  
the \textcolor{black}{\bf{\textit{black}}} vertical and horizontal axes represent the start and the \REVISION{duration} indices while the \textcolor{gray}{\bf{\textit{gray}}} axes represent the corresponding start timestamp and \REVISION{duration} of a moment.
The values in the 2D map, highlighted by \textcolor{red}{\bf{\textit{red}}} color, indicate the matching scores between the moment candidates and the target moment.
Here, $\tau$ is a predefined short duration. The white boxes in the map indicate invalid moments.
(b) Sparse Multi-Scale 2D Temporal Map: 
the sparse multi-scale 2D temporal map is composed of a series of 2D maps under different time units ($\tau$, $2\tau$ and $4\tau$ in this figure). The color of boxes and axes follow the previous definition. In addition, the \textcolor{gray}{\bf{\textit{gray}}} boxes on the 2D map represent the valid moments that are not selected.
In this configuration, we are able to reduce the computational cost by modeling on smaller maps. 
% \HW{Advices: ->} %需要修改：1- (a) (b) 两个标题区分左右两幅图, 放在底部。2- 左边有点挤，右边的空白多。3- 左边图上下没有线条指示。 4- 红色还是黄色，颜色对不上
}
\label{fig:task}
\end{figure*}

\IEEEPARstart{T}{emporal} localization is a fundamental problem in video content understanding. There are several related tasks, such as temporal action localization~\cite{SSN2017ICCV,chen2019relation,zeng2019breaking,xu2020g,lin2021learning}, anomaly detection~\cite{hasan2016learning}, video summarization~\cite{song2015tvsum,chu2015video}, and moment localization with natural language~\cite{gao2017tall,hendricks17iccv,escorcia2019temporal,zhao2021cascaded,lin2020moment}. 
%Several related tasks are proposed for different scenarios, such as temporal action localization~\cite{SSN2017ICCV}, anomaly detection~\cite{hasan2016learning}, video summarization~\cite{song2015tvsum,chu2015video}, and moment localization with natural language~\cite{gao2017tall,hendricks17iccv}.
Among them, moment localization with natural language is the most challenging one due to the complexity of moment description.
This task is introduced recently by Gao \textit{et al.} and Hendricks \textit{et al.}~\cite{hendricks17iccv,gao2017tall}.
It aims to retrieve a temporary segment from an untrimmed video, as queried by a given natural language sentence. 
For example, given a query ``\textit{a guy is playing the saxophone}'' and a paired video, the task is to localize the best matching moment described by the query, as shown in Fig.~\ref{fig:task}~(Query A). Video moment localization with natural language has a wide range of applications, such as video question answering~\cite{lei2018tvqa}, video content retrieval~\cite{Shao_2018_ECCV}, dense video captioning~\cite{duan2018weakly}, as well as video storytelling~\cite{gella2018dataset}. 

Most current language-queried moment localization methods follow a two-step pipeline~\cite{gao2017tall,hendricks17iccv,Ge_2019_WACV,liu2018attentive,song2018val}. It first utilizes sliding windows to generate moment segments from the input videos. Next, each moment is matched with the query sentence to estimate whether it is the target moment. 
% This pipeline considers different moment candidates separately and neglects their temporal contexts. 
% Thus it is difficult to model a moment that occurs in the context of other moments and estimate the precise time boundary of the moment. 
This pipeline models different moments independently and neglects their temporal contexts.
Thus it is difficult to precisely localize a moment that requires context information from other moments.
For example, as shown in Fig.~\ref{fig:task}~(Query C), it targets to localize the query ``\textit{the guy plays the saxophone \emph{\textbf{again}}}'' in the video. If the model only watches the temporal moments in the latter part of the video, it cannot localize the described ``\textit{\emph{\textbf{again}}}'' moment accurately.
Moreover, as shown in Fig.~\ref{fig:task}~(Query B), there are many temporal candidates overlapping with the target moment (the visualized time slots). These candidates are relevant with respect to visual content, but they depict different semantics. 
%Previous methods process each moment candidate individually, therefore, it is difficult for them to distinguish visually similar moments. 
%It is difficult to distinguish these visually similar moments, because that previous methods process each moment candidate individually. 
Consequently, it may be difficult for existing methods to distinguish these visually similar moments since they process each temporal candidate individually.
%It is difficult for previous methods to distinguish these visually similar moments since they process each moment candidate separately.

To address these critical problems in moment localization with natural language, we propose a novel method, 2D Temporal \REVISION{Adjacency} Networks (2D-TAN). The core idea is to localize a target moment from a two-dimensional temporal map, instead of a conventional one-dimensional sequential modeling  as presented in Fig.~\ref{fig:task}. More specifically, for a temporal map with time unit $\tau$, the $(i,j)$\textit{-th} location on it represents a candidate moment starting from the timestamp $i\tau$ and lasting for $(j+1)\tau$. This kind of 2D temporal map covers various video moments with different lengths, while representing their adjacent relations. In this fashion, 2D-TAN can perceive more moment context information when predicting whether a moment is related to other temporal segments. On the other hand, the adjacent moments in the map have content overlap but may depict different semantics. Considering them as a whole, 2D-TAN is able to learn discriminative features to distinguish them.

The time unit $\tau$ on the 2D temporal map decides the localization granularity. To guarantee a fine-grained localization of action instances, especially for precise temporal boundaries, we further build up multi-scale 2D temporal maps, in which multiple choices of time unit are considered. %We construct $K$ 2D temporal maps with different time units and maximum duration. 
We construct $K$ 2D temporal maps, each of which enumerates possible moments based on its time unit and longest duration.
% with different time units and maximum duration. 
% \textcolor{red}{The $k$-th map ($0\el k \le K-1$) enumerates all possible moments shorter than $2^kA\tau$ with the time unit of $2^k\tau$, where $A$ is a constant ratio between the moments' longest duration and the shortest duration for all temporal maps.}
%For each map, we enumerate all possible moments based on the its time unit and longest duration.
% For each map, we enumerate all possible moments based on the its time unit and longest duration.
% Different from the enumeration of dense proposal candidates,  
%Such multi-scale modeling allows the localization networks to perceive moment candidates with more diverse temporal ranges. The multi-scale receptive fields enable the networks to obtain richer context during inference.
Such multi-scale modeling allows the localization networks to perceive moment candidates with more diverse temporal ranges and richer context.
%In this way, moments can obtain larger receptive field and leverage richer context from different scales. 
% To reduce the computation cost in multi-scale temporal maps, we perform sparsely sampling on the dense temporal map.
% Moreover, the multi-scale modeling can be regarded as a sparse sampling with different intervals on the dense single-scale temporal map, thus could reduce the computation cost of moment generation. %it does not increase much computation cost for moment generation. %
% Compared with the dense single-scale case, sparse sampling reduces the total number of moment candidates from $O(N^2)$ to $O(N)$ (see section~\ref{sec:memory_speed} for detailed proof), keeping a linear complexity in multi-scale modeling. 
% Therefore the sparse multi-scale map let the model to learn more discriminative features compared to dense single-scale map.
\REVISION{Moreover, the multi-scale maps can be regarded as a sparse sampling with different intervals on a dense single-scale temporal map, thus reducing the computation costs of moment feature extraction and temporal context modeling.
Compared with the dense single-scale modeling, sparse sampling reduces the total number of moment candidates from $O(N^2)$ to $O(N)$. Modeling context on multiple smaller maps, instead of a single large one, can further reduce the complexity from quadratic to linear (see Sec.~\ref{sec:memory_speed} for detailed proof).}

Some preliminary ideas of this paper have appeared in our earlier work~\cite{2DTAN_2020_AAAI}.
In this paper, we extend the previously proposed single-scale 2D moment localization method to a multi-scale version. The new model considers the adjacent relations of moments at different temporal scales while achieving better performance with faster speed and less memory consumption.
The main contributions of this paper are summarized as following:
%two-fold used here is not correct.

\begin{itemize}
    \item We introduce a novel \textit{two-dimensional temporal map} for modeling the temporal adjacent relations of video moments. Compared with previous methods, the 2D temporal map enables the model to perceive more video context information and learn discriminative features to distinguish the moments with complex semantics.
    \item We propose a \textit{Multi-Scale 2D Temporal \REVISION{Adjacency} Network}, \textit{i.e.}, MS-2D-TAN, for moment localization with natural language. 
    The multi-scale modeling allows our model to have a larger receptive field and obtain richer context. Meanwhile, it reduces the complexity of generating moments from quadratic to linear, which makes the dense video prediction more efficient.
    Without any sophisticated video-language cross-modality fusion, MS-2D-TAN achieves competitive performance in comparison with the state-of-the-art methods on three benchmark datasets. Our code and models are publicly available at~\url{https://github.com/microsoft/2D-TAN}.
\end{itemize}

\section{Related Work}

There are two major sub-fields in the temporal localization in untrimmed videos: temporal action localization and moment localization with natural language. Temporal action localization aims to predict the start and the end time and the category of an activity instance in untrimmed videos. The representative methods include two-stage temporal detection methods~\cite{SSN2017ICCV} and one-stage single shot methods~\cite{lin2017single}. 
This task is limited to pre-defined simple actions and cannot handle complex events in the real world. Therefore, moment localization with natural language~\cite{gao2017tall,hendricks17iccv} is introduced recently to tackle this problem. 
In the remainder of this section, we first review recent clip-based and moment-based methods, and then discuss their cross-modality alignment. In addition, we introduce some related methods in other tasks and their difference from our preliminary work.

Localizing moments in videos by query sentences is very challenging. It not only needs to understand video content, but also requires to align the semantics between video and language. 
%Video clips are a set of short video segments with fixed duration and interval, while moments are a sequence of video clips.
A video clip is a set of short video segments with fixed duration and interval, while a moment is a sequence of video clips.
Depending on whether the moment features are extracted, we divide current methods into two categories, clip-based methods (not extracted) and moment-based methods (extracted). 
%In the following, we first introduce these two type of methods and then discuss how they align video information with language. We also introduce some related tasks that consider the 1D pairwise relations on the 2D map.

\textit{Clip-based methods. }
%For clip-based methods, the main idea is to align video clip with language and predict the moment scores without extracting moment features. 
The main idea of clip-based methods is to align video clips with language directly and predict matching scores without extracting moment features. 
% Since clip features and moment scores are not one-to-one mapping, we can divide the .
In general, there are three common ways to map a clip to a moment score: anchor-based methods, anchor-free methods, and RL-based methods.
Anchor-based methods define a set of anchors with a fixed length for each clip~\cite{chen2018temporally,zhang2019cross,yuan2019semantic}, while anchor-free methods directly predict the start and the end time through classification~\cite{zhang2020span,yuan2019to} or regression~\cite{liu2018attentive,ghosh2019excl,zeng2020dense,liu2018crossmodal,mun2020local,Rodriguez_2020_WACV,chen2020rethinking,lu2019debug,jiang2019cross}.
RL-based methods model the task as a sequential decision-making problem and solve it by reinforcement learning~\cite{he2019read,wang2019language,Hahn2019tripping,wu2020tree}. 
% \HW{Along with their different prediction steps, these methods also explore different clip-level context modeling.}
%Context information is effective in visual content modeling.
In terms of their context modeling, most approaches~\cite{chen2018temporally,zhang2019cross,yuan2019to,ghosh2019excl,zeng2020dense,Rodriguez_2020_WACV,liu2018attentive} gradually aggregate the context information through a recurrent structure. 
Some approaches~\cite{yuan2019semantic,liu2018crossmodal,he2019read} model surrounding clips as the local context using 1D convolution layers, while other approaches model the entire clip as the global context through self-attention modules~\cite{zhang2020span,mun2020local,lu2019debug,chen2020rethinking}. 
Since clips are the shortest moments, the clip-level context is a subset of moment-level context.
Thus, all these context modeling methods ignore a large part of moment-level context.
% Since the semantic meaning of the given query is corresponding to a moment (a sequence of clips), only modeling context for each clip will ignore the their temporal correlations.
%Recently, self-attention is also been proposed to model the global context with all other clips~\cite{zhang2020span,mun2020local,lu2019debug,chen2020rethinking}.
% Therefore, they lack the ability to model the 
% \HW{Due to the absence of candidate moments and moment-level refinement, they lack the ability to model moment level relationship. 
% In contrast, our focus is to model relationship on selected moments and learn discriminative moment features to boost the localization performance.}
In contrast, our focus is to model contexts on the moment level and to learn discriminative moment features to boost the localization performance.

\textit{Moment-based methods. }
Moment-based methods extract moment-level features and learn a matching function between features and the query. %a mapping between moment features and language matching scores.
Most approaches predict moment scores in one stage, while some approaches follow a cascaded strategy and filter moments by multiple stages~\cite{xu2019multilevel,chen2019semantic}.
%Different from the context modeling in clip-level, 
Different from clip-based methods, existing moment-based methods integrate temporal context in two common ways. 
One way is to use the whole video as the global context. For example, Hendricks \textit{et al.}~\cite{hendricks17iccv} and He \textit{et al.}~\cite{he2019read} concatenate each moment feature with the global video feature~\cite{hendricks17iccv} as the moment representation.
Wang \textit{et al.} concatenate semantic features with the global video feature~\cite{wang2019language}. 
Another way is to use the surrounding clips as the local context for a moment.
Gao \textit{et al.}, Liu \textit{et al.}, Song \textit{et al.} and Ge \textit{et al.} concatenate the moment feature with clip features before and after current clip as its representation~\cite{gao2017tall,liu2018attentive,song2018val,Ge_2019_WACV}.
Since these methods only consider one or two specific  moments, the rich context information from other possible moments is ignored. 
One recent exploration~\cite{zhang2019man} of aggregating the context from other moments is Graph Convolutional Network (GCN). %don't understand
However, one inherent feature of graph is \textit{node permutation invariance}~\cite{dehmamy2019understanding}, switching any two nodes does not change the result. Therefore, the graph modeling ignores the temporal ordering of different moments.
In contrast, our method considers all the neighboring moments as the context and models the moments on a 2D convolution network. The 2D convolution network can naturally preserve the relative position of different moments.
% This design enables the model to learn more discriminative features without losing the relative location information.
This design enables the model to perceive more context information and learn more discriminative features.
% Different from their works, we define a 2D temporal map and learn segments adjacent relations through the stacked 2D convolution layers.

\textit{Video and language cross-modality alignment.}  
The simplest way for cross-modal alignment is to directly multiply or concatenate the clip/moment features with sentence features~\cite{hendricks17iccv,liu2018crossmodal,gao2017tall}.
Recent approaches improve video and language alignment from several different directions, such as cross-modal attention, sentence syntactic modeling, and compositional reasoning. 
%For cross-modal attention, the key idea is to attend relevant video clips/moments or query words from another modality. 
The key idea of cross-modal attention is to attend relevant video clips/moments or query words from another modality. 
Some methods attend relevant video features through words~\cite{liu2018attentive,jiang2019cross}, while most other methods attend both the relevant video features and words via the co-attention module~\cite{liu2018temporal,lu2019debug,chen2018temporally,zhang2019cross,wang2020temporally,Rodriguez_2020_WACV,zhang2019man,liu2018crossmodal,song2018val,jiang2019cross,yuan2019to,zhang2020span}.
For sentence syntactic modeling, Zhang \textit{et al.}~\cite{zhang2019cross} enhance the sentence modeling with the queries' syntactic graph.
For compositional reasoning, Liu \textit{et al.}~\cite{liu2018temporal} explicitly use the compositionality in natural languages for temporal reasoning in videos.
Instead of using the complex attention modules, our proposed MS-2D-TAN model only adopts a simple multiplication operation for visual and language feature fusion.

\textit{Discussion with related methods in other tasks}
The self-attention operation in document modeling enumerate all possible word pairs. This enumeration is similar to our 2D temporal map, where the start and the end time pairs are enumerated.
Due to the enumeration, both tasks face the same problem: memory cost and speed scale up quadratically with the sequence length. % don't understand
% Long sequence modeling is a common problem in both vision and language field. 
% Common solutions include dilated convolutions, sliding windows and down sampling.
Several methods are recently proposed to improve the efficiency in the long document tasks.
Beltagy \textit{et al.}~\cite{Beltagy2020Longformer} combine global attention, dilated convolution and sliding window together to reduce the complexity to linear.
Sukhbaatar \textit{et al.}~\cite{sukhbaatar2019adaptive} design an adaptive attention span, which learns a hard mask to ignore the long-term relations.
% Chao \textit{et al.}~\cite{chao2018rethinking} build a multi-scale architecture with dilated convolutions for modeling the context in long untrimmed videos.
% Yeung \textit{et al.}~\cite{yeung2016end} train an agent to learn what to look next and when to omit the prediction without observing the entire long video.
% Kitaev \textit{et al.}~\cite{kitaev2020reformer} replace dot-product attention with locality-sensitive hashing to reduce the complexity.
% Since the sequence modalities are different (sentences v.s. videos), these methods are not applicable for the video domain. 
In this paper, we focus on reducing the computational cost in the video domain.
Our study is also related to some contemporaneous and recent work~\cite{lin2019bmn,lin2020fast} that apply 2D temporal map to temporal action localization task. 
In contrast to their work, our work focus on moment localization with natural language. Besides, we also propose a multi-scale 2D temporal map to improve the efficiency and the performance of the dense 2D temporal map.

\textit{Difference from our preliminary work}.
Some preliminary ideas in this paper appeared in the conference version~\cite{2DTAN_2020_AAAI}\REVISION{, where 2D-TAN is proposed to learn discriminative moment features through a 2D temporal map. A sparse single-scale map is  introduced to reduce the computational cost of feature extraction from quadratic to linear. However, the complexity of the 2D temporal adjacent network remains quadratic.
Compared with~\cite{2DTAN_2020_AAAI}, the proposed MS-2D-TAN further reduce the complexity of the 2D temporal adjacent network from quadratic to linear.}
The experimental results (Sec.~\ref{sec:experiments}) show clearly that the new model is more efficient (Fig.~\ref{fig:speed_memory_self} \REVISION{and Fig.~\ref{fig:speed_memory_others}}) and achieves better performance in three benchmark datasets (Table.~\ref{tab:charades}, \ref{tab:activitynet}, and \ref{tab:tacos}).

\section{Our Approach}

%In this section, we will first briefly introduce the formulation of the moment localization with natural language task. Next, we will go through the pipeline of our network: language representation, video representation, multi-scale gated 2D temporal adjacent network and the loss function, as shown in Figure~\ref{fig:framework}.
In this section, we first introduce the problem formulation of moment localization with natural language. 
% \HW{Next, we will go through the pipeline of our network: language representation, video representation, multi-scale gated 2D temporal adjacent network and the loss function, as shown in Fig.~\ref{fig:framework}.}
Next, we present the pipeline of our proposed network, including language representation, video representation and multi-scale 2D temporal \REVISION{adjacency} network, as shown in Fig.~\ref{fig:framework}. Finally, we present the training and the inference of our network.

\subsection{Problem Formulation} 
Given an untrimmed video $V$ and a sentence $S$ as a query, the task is to retrieve the best matching temporary moment $M$ specified by the query. More specifically, we denote the query sentence as $S=\{s_i\}_{i=0}^{l^S-1}$, where $s_i$ represents a single word, and ${l^S}$ is the total number of the words. The input video is a frame sequence, \textit{i.e.} $V=\{x_i\}_{i=0}^{l^V-1}$, where $x_i$ represents a frame in a video and $l^V$ is the total number of frames. 
The retrieved moment starting from frame $x_i$ to $x_j$ delivers the same semantic meaning as the query $S$.

\subsection{Language Representation via Sequential Embedding}

We first extract the feature of an input query sentence. For each word $s_i$ in the input sentence $S$, we generate its embedding vector ${\bf{w}}_i \in \mathbb{R}^{d^S}$ by the GloVe word2vec model~\cite{pennington2014glove}, where $d^S$ is the vector length. We then sequentially feed the word embeddings $\{ {\bf{w}}_i \}_{i=0}^{l^S-1}$ into a three-layer bidirectional LSTM network~\cite{hochreiter1997long}, and use its average output as the feature representation of the input sentence, \textit{i.e.} ${\bf{f}}^S\in \mathbb{R}^{d^{S}}$. The extracted feature encodes the language structure of the query sentence, thus describes the moment of interest.

\subsection{Video Representation via 2D Temporal Feature Map}
\label{sec:MS-2D-Feature-Map}

In this section, we extract moment features from the input video stream and then construct a 2D temporal feature map.

Given an input video, we first segment it into small non-overlapped video clips, where each clip consists of $T$ contiguous frames. 
% Since videos are variable in length, during training, we perform sliding window to randomly select $N$ consecutive video clips.
For each video clip, we extract its feature using pre-trained CNN model (see experiments section for details).
To generate more compressed video clip representation in the channel dimension, we then feed the video clip feature into a fully connected layer with $d^V$ output channels. 
Therefore, the final representation of compressed video clips is represented as $\{\REVISION{\mathbf{f}^V_i}\}_{i=0}^{N-1}$, where $d^V$ is the number of output channels and $N$ is the total number of video clips.

\begin{figure*}
    \centering
    \includegraphics[width=\textwidth]{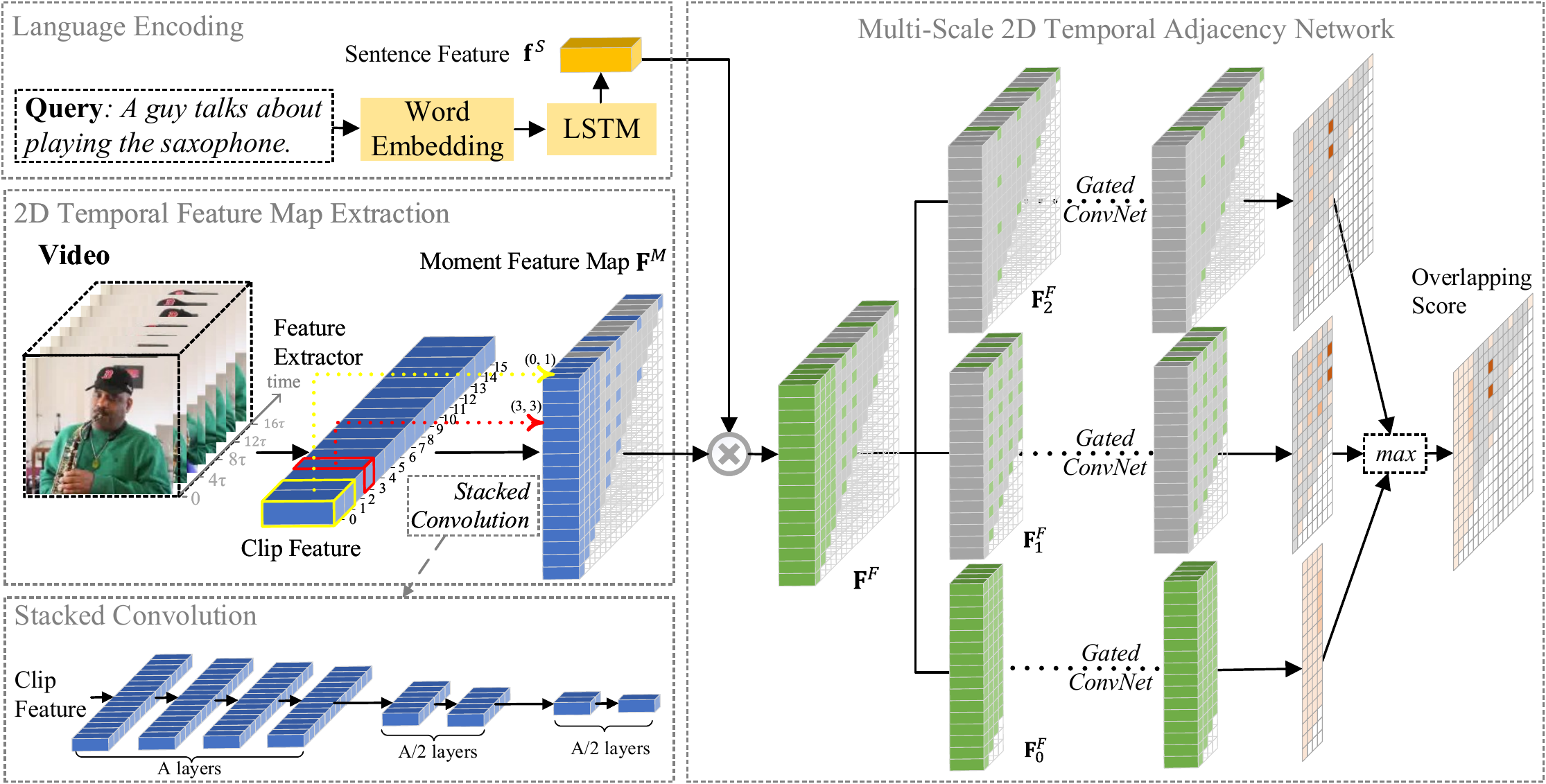}
    \caption{Our proposed framework. The framework is composed of three modules: the language encoding module, the 2D temporal feature map extraction module, and the multi-scale 2D temporal \REVISION{adjacency} network.}
    \label{fig:framework}
\end{figure*}

The $N$ video clips serve as the basic elements for moment candidate construction. 
Thus, we build up feature maps of moment candidates by the video clip features $ {{\{{\bf{f}}^V_i}\}_{i=0}^{N-1}}$.
%In more details, the $k$-th map ($0 \le k \le K-1$) enumerates all possible moments shorter than $2^kA\tau$ with the time unit of $2^k\tau$, where $A$ is number of anchors. For simplicity, we set the same value of $A$ for all temporal maps, which is equal to the ratio between the longest duration and the shortest duration for each temporal map.
Previous works extract moment features from clip features in two ways: pooling~\cite{hendricks17iccv} or stacked convolution~\cite{zhang2019man}.
In this work, we follow the stacked convolution design. 
In the stacked convolution, the output of each layer are moment features, where these moments have equal length but start at different time, as shown in Fig.~\ref{fig:framework}.
%We can collect all possible moment features by stacking $N-1$ layers of kernel size $2$ and stride $1$. However, this become prohibitive when $N$ is large.
In order to generate moment features, a simple way is to stack $N$ convolution layers, where each layer has a kernel of size $2$ and stride $1$ (except for the first layer, which has kernel size $1$ and stride $1$).
 However, when $N$ is large, it is infeasible to fit the data into memory.
Instead of enumerating all of them, we conduct a sparse sampling strategy as following:
%In more details, there are $\frac{(K+1)A}{2}$ convolution layers in total.
We first define two constant number $A$ and $K$.
For the first convolution layer, we set both the kernel size and stride to $1$.
For the $(i+1)\frac{A}{2}$-th layers $(1\le i\le K-1)$, the kernel size and stride are set to $3$ and $2$, respectively.
All the rest layers have kernels of size $2$ and strides $1$. 
We also add batch norm after each convolution layer and use $Tanh$ as the activation function.
Therefore, there are $\frac{(K+1)A}{2}$ convolution layers in total. 
% Comparing with enumeration, the sparse sampling strategy is more efficient when $N$ is large.
In this way, we densely sample moments of short duration, and gradually increase the sampling interval when the moment duration becomes long.
%In this way, features for long-time moment candidates are extracted at higher convolution layers, while features for s are extracted at lower convolution layers.
In more details, when the number of sampled clips is small, \textit{i.e.} $N\le A$, we enumerate all possible moments as candidates. 
When $N$ becomes large, \textit{i.e.} $N>A$, a moment starting from clip $v_a$ to $v_b$ is selected as the candidate when satisfying the following condition $G(a, b)$:
\begin{equation}
       G(a,b) \Leftarrow 
             (a~\emph{mod}~s {\tiny{=}} 0) ~~\&~~ (b~\emph{mod}~s {\small{=}} 0),
\label{eq:candidate1}
\end{equation}
where $a$ and $b$ are the indexes of clips, $s$ is defined as $s=2^{k-1}$, $k=\lceil\log_2(\frac{b-a+1}{A})+1\rceil$ and $\lceil \cdot \rceil$ is the ceiling function.  
If $G(a,b)=1$, the moment is selected as the candidate, otherwise, it is not selected.
This sampling strategy can largely reduce the number of moment candidates, as well as the computational cost.

Different from previous methods which directly operate on an individual video moment, we restructure the whole sampled moments to a 2D temporal feature map, denoted as $\mathbf{F}^M\in \mathbb{R}^{N\times N\times d^V}$. 
The 2D temporal feature map $\mathbf{F}^M$ consists of three dimensions: the first two dimensions $N$ represent the start and \REVISION{duration} clip indexes respectively, while the third one $d^V$ indicates the feature dimension.
The feature of a moment starting from clip $v_a$ and durating $b$ clips is located at $\mathbf{F}^M[a,b,:]$ on the feature map.
Denoting the $i$-th output at $j$-th convolution layer as $\mathbf{f}^M[i,j]\in \mathbb{R}^{d^V}$, it corresponds to the $(a,b)$-th location on the feature map $\mathbf{F}^M$, where 

\begin{equation}
    \begin{aligned}
    & a = 2^ki,\\
    & b = 
    \begin{cases}
    	j & \text{if $j<A$,}\\
    	A + 2^{\lceil \frac{2(j-A+1)}{A}\rceil}\cdot (j-A+1)-1 & \text{otherwise.}\\
    \end{cases}
    \end{aligned}
\label{eq:multi_scale_map}
\end{equation}

Noted that, the moment's start and duration clip indexes $a$ and $b$ should satisfy $a+b<N$.
Therefore, on the 2D temporal feature map, all the moment candidates locating at the region of $a+b\ge N$ are invalid, \textit{i.e.} the lower triangular part of the map, as shown in Figure~\ref{fig:framework}. The values in this region are padded with zeros in implementation.

\subsection{Multi-Scale 2D Temporal \REVISION{Adjacency} Network}

After obtaining the language and video features, we then predict the best matching moment queried by the sentence from all candidates.

We first fuse the 2D temporal feature maps $\textbf{F}^M$ with the sentence feature $\textbf{f}^S$.
Specifically, we project these two cross-domain features to the same subspace via fully connected layers, and then fuse them through Hadamard product and $\ell_2$ normalization as

\begin{equation}
    \mathbf{F}^F = \| ({\bf{w}}^S \cdot {\mathbf{f}}^S \cdot {\mathbbm{1}}^T) \odot ({\bf{W}}^M \cdot {\mathbf{F}}^M) \|_F,
\label{eq:fusion}
\end{equation}

where ${\bf{w}}^S$ and ${\bf{W}}^M$ represent the learnt parameters of the fully connected layers, ${\mathbbm{1}}^T$ is the transpose of an all-ones vector, $\odot$ is Hadamard product, and $\|\cdot\|_F$ denotes Frobenius normalization.
The fused feature map is denoted as $\mathbf{F}^F \in \mathbb{R}^{N\times N\times d^F}$, where $d^F$ is the feature dimension after fusion. 

Different from our preliminary work which directly builds up a temporal \REVISION{adjacency} network over the single-scale map $\mathbf{F}^F$,
in this paper, we build up multi-scale temporal \REVISION{adjacency} networks over the sparse multi-scale maps, %\REVISION{which requires less time and memory usage}.
\REVISION{which executes faster and takes lower memory footprints}. In more details, we first restructure the single-scale map $\mathbf{F}^F$ to $K$ multi-scale maps.
%We denote the $i$-th output at $j$-th convolution layer as $\mathbf{f}^M[i,j]\in \mathbb{R}^{d^V}$.
%, denoted as $\{\mathbf{F}^F_k | \mathbf{F}^F_k\in \mathbb{R}^{N\times {2^k}A\times d^F}, 0\le k \le K-1\}$, where $K$ is the total number of scales and $A$ defines the ratio between the longest and shortest moment duration at each scale.
The $k$-th sparse map is sampled from the fused feature map with interval $2^k$.
Specifically, all the $(2^ki-1,2^kj-1)$-th locations on $\mathbf{F}^F$ are sampled, where $1\le i \le N$ and $1\le j \le A$ and $A$ defines the number of anchors at each scale.
In this way, the shortest and longest moments in the $k$-th map are of size $2^k$ and $2^kA$, respectively.
We denote the multi-scale map as $\{\mathbf{F}^F_k | \mathbf{F}^F_k\in \mathbb{R}^{N\times {2^k}A\times d^F}, 0\le k \le K-1\}$.

Next, we build up temporal \REVISION{adjacency} networks over the multi-scale 2D feature maps.
In more details, each map corresponds to $L$ gated convolutional layers~\cite{yu2019free} with kernel size of $\kappa$, dilation size of $2^k$ and stride size of $2^k$.
The output of the $L$ layers keeps the same shape as the input fused feature map through zero padding at each layer.
This design enables the model to gradually perceive more context of adjacent moment candidates, while learn the difference between moment candidates. 
Moreover, the receptive field of the network is large, thus it can observe a large portion of video content, resulting in learning the temporal contexts.
%Then, we adopt the Gated Convolution Layer~\cite{yu2019free} to handling this problem.
%as shown in Formula~\ref{eq:gate_conv}.

%\begin{equation}
%    \begin{aligned}
%    \mathbf{G}^{i}_k[x,y] &= \Sigma \Sigma \mathbf{U}^{i}_k \mathbf{F}^{i}_k[x,y]\\
%    \mathbf{O}^{i}_k[x,y] &= \Sigma \Sigma \mathbf{W}^{i}_k \mathbf{F}^{i}_k[x,y]\\
%    \mathbf{F}^{i+1}_k[x,y]&=\phi(\mathbf{O}^{i}_k[x,y])\odot\sigma(\mathbf{G}^i_k[x,y]) \\
%    \end{aligned}
%\label{eq:gate_conv}
%\end{equation}
%
%
%where $\phi$ is ReLU function, $\sigma$ is sigmoid function, $\mathbf{F}^{i}_k$, $\mathbf{G}^{i}_k$, $\mathbf{O}^{i}_k$, $\mathbf{F}^{i+1}_k$, $\mathbf{U}^i_k$ and $\mathbf{W}^i_k$, are the input feature, gate, output of convolution, the output feature after masking, the convolution kernel of gate and the convolution kernel of output for the $i$-th convolution layer of $k$-th 2D temporal feature map.
% Additionally, we also find that the kernel size of TAN has strong influence on the final performance. 
% We can also further reduce the computation cost by applying the Separable Convolution design proposed by Peng \textit{et al.}~\cite{peng2017large}.

Finally, we predict the matching scores of moment candidates with the given sentence on the multi-scale 2D temporal maps. We feed the output feature maps into fully connected layers and the $sigmoid$ function separately, then generate multi-scale 2D score maps.
The valid scores on each score map are then collected, denoted as $P_k=\{p_k^i\}_{i=1}^{C_k}$, where $C_k$ is the total number of valid moment candidates of the $k$-th map.
Each value $p_k^i$ on the $k$-th map represents the matching score of a moment candidate with the sentence. 

\subsection{Training and Inference}

During training, we adopt a scaled $IoU$ value as the supervision signal, rather than a hard binary score. 
Specifically, for each moment candidate, we compute its $IoU$ $o_k^i$ with the ground truth moment. The $IoU$ score $o_k^i$ is then scaled with thresholds $0.5$ as
\begin{equation}
\begin{aligned}
    y_k^i& =
    \begin{cases}
    0 & {{o_k^i} \le 0.5}, \\
    2o_k^i-1 & o_k^i > 0.5, \\
    \end{cases}
\label{eq:iou}
\end{aligned}
\end{equation}
and $y_k^i$ serves as the supervision label. 
Our network is trained by a binary cross entropy loss as
\begin{equation}
    Loss =\sum_{k=0}^{K-1}\sum_{i=1}^{C_k} y_k^i \log p_k^i + (1-y_k^i) \log(1-p_k^i),
\label{eq:loss}
\end{equation}
where $K$ is the total number of maps, $p_k^i$ and $C_k$ are the output score of a moment and
the total number of valid candidates on the $k$-th map.

During inference, the score maps $\{P_k,0\le k\le K-1\}$ are recovered to a single-scale map $P'$ based on the moment location at the original single-scale map. Noted that some moments are predicted by more than one score map. And we choose the highest score as its final prediction.

\section{Experiments}
\label{sec:experiments}

In this section, we first introduce the datasets used in our experiments and then go through the experiment settings. Next, we compare our proposed MS-2D-TAN with previous work. Finally, we conduct ablation studies and visualize experimental results.

\subsection{Datasets}

\begin{figure*}[t]
    \centering
    \includegraphics[width=\textwidth]{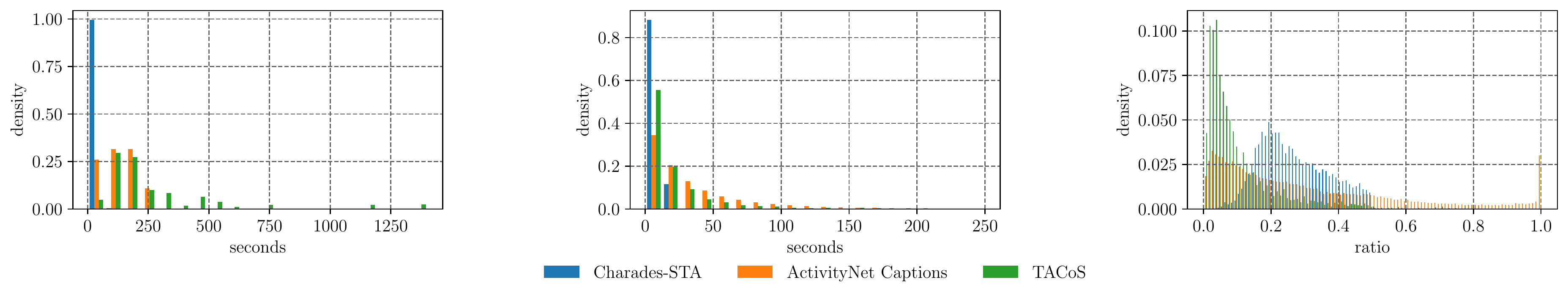}
    \caption{Comparing Charades-STA, ActivityNet Captions and TACoS. \textbf{Left:} comparing video duration. \textbf{Middle:} comparing target moment duration. \textbf{Right:} comparing the ratio between target moment duration and video duration.}
    \label{fig:distribution}
\end{figure*}

We evaluate our methods on the following three datasets:

\textbf{Charades-STA~\cite{gao2017tall}.} It contains $9,848$ videos of daily indoor activities. It is originally designed for action recognition and localization. Gao \textit{et al.}~\cite{gao2017tall} extend the temporal annotation (\textit{i.e.} the start and end time of moments) of this dataset with language descriptions. Charades-STA contains $12,408$ moment-sentence pairs in training set and $3,720$ pairs in testing set.

\textbf{ActivityNet Captions~\cite{krishna2017dense}.} It consists of $19,209$  videos, whose content are diverse and open.
It is originally designed for dense video captioning, and recently introduced to moment localization with natural language, since these two tasks are reversible~\cite{chen2018temporally,zhang2019cross}.
Following the experimental setting in~\cite{zhang2019cross}, we use val\_1 as the validation set and val\_2 as the testing set , which have $37,417$, $17,505$, and $17,031$ moment-sentence pairs for training, validation, and testing, respectively. Currently, this is the largest dataset in this task.

\textbf{TACoS~\cite{regneri2013grounding}.}
It consists of 127 videos selected from the MPII Cooking Composite Activities video corpus~\cite{rohrbach2012script}, which contain different activities happened in the kitchen room.
Regneri \textit{et al.} extend the sentence descriptions by crowd-sourcing.
A standard split~\cite{gao2017tall} consists of $10,146$, $4,589$, and $4,083$ moment-sentence pairs for training, validation and testing, respectively.

We also demonstrate the distribution of video duration, target moment duration and their ratio of these three datasets in Fig.~\ref{fig:distribution}.
We can observe that videos in TACoS dataset have more variant and longer duration than others (see Fig.~\ref{fig:distribution} Left).
Charades-STA has shorter target moments compared to ActivityNet Captions and TACoS (see~\ref{fig:distribution} Middle).
By comparing the ratio between the target moment and the entire video (see Fig.~\ref{fig:distribution} Right), we find that most target moments in TACoS have smaller ratios.
% In order to precisely cover most target moments in our 2D temporal map, We need larger $N$ for TACoS than others.

\subsection{Evaluation Metric}
Following the setting in previous work~\cite{gao2017tall}, we evaluate our model by computing $Rank$ $n$@$m$. It is defined as the percentage of language queries having at least one correct moment retrieval in the top-$n$ retrieved moments. A retrieved moment is correct when its IoU with the ground truth moment is larger than $m$. 
There are specific settings of $n$ and $m$ for different datasets. Specifically,
we report the results as $n\in\{1, 5\}$ with $m\in\{0.5,0.7\}$ for Charades-STA dataset, $n\in\{1, 5\}$ with $m\in\{0.3,0.5,0.7\}$ for ActivityNet Captions dataset, and $n\in\{1, 5\}$ with $m\in\{0.1,0.3,0.5,0.7\}$ for TACoS dataset.

\subsection{Implementation Details}
The performance of video features plays an important role in moment localization with natural language.
To make a thorough and fair comparison, we collect the most commonly used features in previous works, including VGG~\cite{simonyan2014very}, C3D~\cite{tran2015learning} and I3D~\cite{carreira2017quo} features.
% In general, C3D features~\cite{tran2015learning} and I3D features~\cite{carreira2017quo} are most commonly used in all three datasets. VGG features~\cite{simonyan2014very} are frequently used in some earlier works on Charades-STA and TACoS datasets. Most recently, some work also report their results on Charades-STA with I3D features finetuned on Charades~\cite{sigurdsson2016hollywood}. 
The details about video feature extractor are illustrated as follows:
\begin{itemize}[leftmargin=0.38cm]
\item{
\textbf{VGG}. Following~\cite{zhang2019man}, we use VGG$16$ pre-trained on ImageNet~\cite{deng2009imagenet}. Specifically, videos are decoded at $24$ fps and  the output of $fc7$ layer after $ReLU$ are extracted at $6$ fps. The clip feature corresponds to every $4$ consecutive frame.
Therefore, each clip corresponds to $1$ second.
}
\item{
\textbf{C3D}. Following~\cite{wang2020temporally}, we use C3D network pre-trained on sport1M~\cite{karpathy2014large}. Specifically, videos are decoded at $16$ fps and the output of $fc6$ layer after ReLU are extracted for every $16$ consecutive frames. 
Each video clip corresponds to $1$ second.
}
\item{
\textbf{I3D}. Following~\cite{yuan2019semantic}, we use I3D network pre-trained on Kinetics~\cite{carreira2017quo}. 
Specifically, videos are decoded at $25$ fps and the output of the last average pooling layer are extracted for every $16$ consecutive frames.
Therefore, each video clip corresponds to $0.64$ second.
For Charades-STA, we also finetune the I3D on the Charades dataset~\cite{sigurdsson2016hollywood}.
}
\end{itemize}

Some predicted moments highly overlap with each other. To reduce the redundant prediction, we adopt non-maximum supression (NMS) based on the predicted scores $P'$. We fix the IoU threshold for NMS to $0.49$ for all the experiments. After NMS, we use the top-$n$ ranked moments for evaluation. Please note that NMS does not affect the top-$1$ result.

During training, we adopt sliding window to random select $N$ consecutive clips .
% If the whole video are less than $N$, we pad the clip features to the window size with zeros.
\REVISION{
For a fair comparison, we set the number of hidden states $H=512$, the window size $N=64$, the number of scales $K=3$, the number of anchors $A=16$, the convolution kernel size $\kappa=17$ and the number of layers $L=2$ for all three datasets as default.} 
% On Charades-STA and TACoS, we set $N$ to $64$ and $512$ respectively.
% For ActivityNet Captions, we rescale each feature sequence to $128$ and set $N$ to $128$, following~\cite{mun2020local}.
% For simplicity, we use the same hyper parameters for all maps (\textit{i.e.}  number of anchors, convolution layers and their kernel sizes). 
% We also list the hyper parameters with the best performance for the three datasets, as shown in Table~\ref{tab:setting}.
All these anchors and scale settings is able to cover at least \REVISION{$95\%$ target moments of the training set} with an IoU threshold of $0.7$. 
\REVISION{Finetuning these hyperparameters on specific dataset and feature type could get better performance as demonstrated in Table~\ref{tab:tacos} (the results of MS-2D-TAN$^\star$), where $H=512$, $N=512$, $A=8$, $\kappa=9$, and $L=2$.}
To train MS-2D-TAN from scratch, the learning rate is set to $0.0001$ without weight decay. The batch size is set to $32$ and Adam~\cite{kingma2014adam} is used as the optimizer.
% The  hidden state sizes of all layers in our model are set to $512$.

% \begin{table}[t]
%     \caption{Network settings of different datasets.}
%     \centering
%     \begin{tabular}{|c|c|c|c|}
%     \hline
%          & Charades-STA & ActivityNet Captions & TACoS \\
%          \hline
%          $N$ & $64$ & $128$ & $512$ \\
%          $A$ & $16$ & $16$ & $8$ \\
%          $L$ & $2$ & $4$ & $2$ \\
%          $\kappa$ & $17$ & $9$ & $9$ \\
%          $K$ & $3$ & $4$ & $5$ \\
%     \hline
%     \end{tabular}
%     \label{tab:setting}
% \end{table}

\subsection{Comparison with State-of-the-Art Methods}

\begin{table}[t]
	\caption{Performance Comparison on Charades-STA.}
	\begin{center}
		\begin{tabular}{|c|c|c|c|c|c|}
			\hline
			\multicolumn{2}{|c|}{\multirow{2}*{Method}} & \multicolumn{2}{c|}{$Rank1@$} & \multicolumn{2}{c|}{$Rank5@$}  \\
			\cline{3-6}
			\multicolumn{2}{|c|}{} & $0.5$ & $0.7$ & $0.5$ & $0.7$ \\
			\hline
			\multicolumn{6}{|c|}{VGG features} \\
			\hline
			\multicolumn{2}{|@{}c@{}|}{MCN~\cite{hendricks17iccv}} & $17.46$ & $8.01$ & $48.22$ & $26.73$ \\
			\multicolumn{2}{|c|}{MAN~\cite{zhang2019man}}  & $\mathit{41.24}$ & $\mathit{20.54}$ & $\mathit{83.21}$ & $51.85$ \\
			\multicolumn{2}{|c|}{SM-RL~\cite{wang2019language}}& $24.36$ & $11.17$ & $61.25$ & $32.08$ \\
			\multicolumn{2}{|c|}{SAP~\cite{chen2019semantic}} & $27.42$ & $13.36$ & $66.37$ & $38.15$ \\
		  %  \multicolumn{2}{|c|}{DRN~\cite{zeng2020dense}} & $42.90$ & $23.68$ & $87.80$ & $54.87$ \\
		  %  \multicolumn{2}{|c|}{DPIN~\cite{wang2020dual}} & $47.98$ & $26.96$ & $85.53$ & $55.00$ \\
			\hline
			\multicolumn{2}{|c|}{\textbf{MS-2D-TAN}} & $\mathbf{45.65}$ & $\mathbf{27.20}$ & $\mathbf{86.72}$ & $\mathbf{56.42}$ \\
			\hline
			\hline
			\multicolumn{6}{|c|}{C3D features} \\
			\hline
			\multicolumn{2}{|c|}{CTRL~\cite{gao2017tall}} & $23.63$ & $8.89$ & $58.92$ & $29.52$ \\
			\multicolumn{2}{|c|}{ACRN~\cite{liu2018attentive}} & $20.26$ & $7.64$ & $71.99$ & $27.79$ \\
			\multicolumn{2}{|c|}{ROLE~\cite{liu2018crossmodal}} & $21.74$ & $7.82$ & $70.37$ & $30.06$ \\
			\multicolumn{2}{|c|}{VAL~\cite{song2018val}}& $23.12$ & $9.16$ & $61.26$ & $27.98$ \\
			\multicolumn{2}{|c|}{ACL-K~\cite{Ge_2019_WACV}} & $30.48$ & $12.20$ & $64.84$ & $35.13$ \\
			\multicolumn{2}{|c|}{DEBUG~\cite{lu2019debug}}  & $37.39$ & $17.69$ & $-$ & $-$ \\
			\multicolumn{2}{|c|}{GDP~\cite{chen2020rethinking}}  & $\mathit{39.47}$ & $18.49$ & $-$ & $-$ \\
			\multicolumn{2}{|c|}{RWM-RL~\cite{he2019read}} & $36.70$ & $-$ & $-$ & $-$ \\
			\multicolumn{2}{|c|}{QSPN~\cite{xu2019multilevel}} & $35.60$ & $15.80$ & $79.40$ & $45.40$ \\
			\multicolumn{2}{|c|}{SLTA~\cite{jiang2019cross}} & $22.81$ & $8.25$ & $72.39$ & $31.46$ \\
			\multicolumn{2}{|c|}{ABLR~\cite{yuan2019to}}  & $24.36$ & $9.01$ & $-$ & $-$  \\
 			\multicolumn{2}{|c|}{TripNet~\cite{Hahn2019tripping}} & $36.61$ & $14.50$ & $-$ & $-$ \\
			\multicolumn{2}{|c|}{CBP~\cite{wang2020temporally}} & $36.80$ & $\mathit{18.87}$ & $70.94$ & $ \mathbf{50.19}$ \\
		    \multicolumn{2}{|c|}{TSP-PRL~\cite{wu2020tree}} & $37.39$ & $17.69$ & $-$ & $-$ \\
		  %  \multicolumn{2}{|c|}{DRN~\cite{zeng2020dense}} & $45.40$ & $26.40$ & $88.01$ & $55.38$ \\
		    \hline
			\multicolumn{2}{|c|}{\textbf{MS-2D-TAN}} & $\textbf{41.10}$ & $\textbf{23.25}$ & $\mathbf{81.53}$ & $\textit{48.55}$ \\
			\hline
    		\hline
    		\multicolumn{6}{|c|}{I3D features} \\
    		\hline
			\multicolumn{2}{|c|}{SCDM~\cite{yuan2019semantic}} & $\mathit{54.44}$ & $\mathit{33.43}$ & $74.43$ & $\mathit{58.08}$ \\
		    \multicolumn{2}{|c|}{DRN~\cite{zeng2020dense}} & $53.09$ & $31.75$ & $\mathit{89.06}$ & $\mathit{60.05}$ \\
			\hline
			\multicolumn{2}{|c|}{\textbf{MS-2D-TAN}} & $\mathbf{56.64}$ & $\mathbf{36.21}$ & $\mathbf{89.14}$ & $\mathbf{61.13}$ \\
    		\hline
    		\hline
    		\multicolumn{6}{|c|}{I3D features finetuned on Charades} \\
    		\hline
			\multicolumn{2}{|c|}{ExCL~\cite{ghosh2019excl}} & $44.10$ & $22.40$ & $-$ & $-$ \\    	
			\multicolumn{2}{|c|}{TMLGA~\cite{Rodriguez_2020_WACV}} & $52.02$ & $33.74$ & $-$ & $-$ \\
			\multicolumn{2}{|c|}{LGI~\cite{mun2020local}} & $\mathit{59.46}$ & $\mathit{35.48}$ & $-$ & $-$ \\
			\multicolumn{2}{|c|}{VSLNet~\cite{zhang2020span}} & $54.19$ & $35.22$ & $-$ & $-$ \\
			\hline
			\multicolumn{2}{|c|}{\textbf{MS-2D-TAN}} & $\mathbf{60.08}$ & $\mathbf{37.39}$ & $\mathbf{89.06}$ & $\mathbf{59.17}$ \\
			\hline
		\end{tabular}
	\end{center}
	\label{tab:charades}
\end{table}

\begin{table}[t]
	\caption{Performance comparison on ActivityNet Captions.}
\setlength{\tabcolsep}{1.0pt}
	\begin{center}
		\begin{tabular}{|c|c|c|c|c|c|c|c|}
			\hline
			\multicolumn{2}{|c|}{\multirow{2}*{Method}} & \multicolumn{3}{c|}{$Rank1@$} &
			\multicolumn{3}{c|}{$Rank5@$} \\
			\cline{3-8}
			\multicolumn{2}{|c|}{} & $0.3$ & $0.5$ & $0.7$ & $0.3$ & $0.5$ & $0.7$ \\
			\hline
			\multicolumn{8}{|c|}{C3D features} \\
			\hline
			\multicolumn{2}{|@{}c@{}|}{MCN~\cite{hendricks17iccv}} & $39.35$ & $21.36$ & $6.43$ & $68.12$ & $53.23$ & $29.70$ \\
			\multicolumn{2}{|c|}{CTRL~\cite{gao2017tall}} & $47.43$ & $29.01$ & $10.34$ & $75.32$ & $59.17$ & $37.54$ \\
			\multicolumn{2}{|c|}{TGN~\cite{chen2018temporally}} & $43.81$ & $27.93$ & $-$ & $54.56$ & $44.20$ & $-$ \\
			\multicolumn{2}{|c|}{ACRN~\cite{liu2018attentive}} & $49.70$ & $31.67$ & $11.25$ & $76.50$ & $60.34$ & $38.57$ \\
		  %  \multicolumn{2}{|c|}{BSSTL~\cite{li2019bidirectional}} & $55.32$ & $47.68$ & $-$ & $70.53$ & $57.53$ & $-$ \\
			\multicolumn{2}{|c|}{DEBUG~\cite{lu2019debug}} & $55.91$ & $39.72$ & $-$ & $-$ & $-$ & $-$ \\
			\multicolumn{2}{|c|}{GDP~\cite{chen2020rethinking}} & $56.17$ & $39.27$ & $-$ & $-$ & $-$ & $-$ \\
			\multicolumn{2}{|c|}{CMIN~\cite{zhang2019cross}} & $\mathbf{63.61}$ & $43.40$ & $23.88$ & $\mathit{80.54}$ & $67.95$ & $\mathit{50.73}$ \\
			\multicolumn{2}{|c|}{RWM-RL~\cite{he2019read}} & $-$ & $36.90$ & $-$ & $-$ & $-$ & $-$ \\
			\multicolumn{2}{|c|}{QSPN~\cite{xu2019multilevel}} & $52.13$ & $33.26$ & $13.43$ & $77.72$ & $62.39$ & $40.78$ \\
			\multicolumn{2}{|c|}{ABLR~\cite{yuan2019to}} & $55.67$ & $36.79$ & $-$ & $-$ & $-$ & $-$ \\
			\multicolumn{2}{|c|}{TripNet~\cite{Hahn2019tripping}} & $48.42$ & $32.19$ & $13.93$ & $-$ & $-$ & $-$ \\
			\multicolumn{2}{|c|}{SCDM~\cite{yuan2019semantic}} & $54.80$ & $36.75$ & $19.86$ & $77.29$ & $64.99$ & $41.53$ \\
			\multicolumn{2}{|c|}{CBP~\cite{wang2020temporally}} & $54.30$ & $35.76$ & $17.80$ & $77.63$ & $65.89$ & $ 46.20$ \\
		    \multicolumn{2}{|c|}{TSP-PRL~\cite{wu2020tree}} & $56.08$ & $38.76$ & $-$ & $-$ & $-$ & $-$ \\
		    \multicolumn{2}{|c|}{DRN~\cite{zeng2020dense}} & $-$ & $\mathit{45.45}$ & $\mathit{24.36}$ & $-$ & $\mathit{77.97}$ & $50.30$ \\
			\multicolumn{2}{|c|}{LGI~\cite{mun2020local}} & $58.52$ & $41.51$ & $23.07$ & $-$ & $-$ & $-$ \\
% 			\multicolumn{2}{|c|}{DPIN~\cite{wang2020dual}} & $62.40$ & $47.27$ & $28.31$ & $87.52$ & $77.45$ & $60.03$ \\
			\hline
			\multicolumn{2}{|c|}{\textbf{MS-2D-TAN}} & $\REVISION{\mathit{61.04}}$ & $\REVISION{\mathbf{46.16}}$ & $\REVISION{\mathbf{29.21}}$ & $\REVISION{\mathbf{87.30}}$ & $\REVISION{\mathbf{78.80}}$ & $\REVISION{\mathbf{60.85}}$ \\
			\hline
			\hline
			\multicolumn{8}{|c|}{I3D features} \\
			\hline
			\multicolumn{2}{|c|}{ExCL~\cite{ghosh2019excl}} & $\mathit{62.30}$ & $42.70$ & $24.10$ & $-$ & $-$ & $-$ \\
			\multicolumn{2}{|c|}{TMLGA~\cite{Rodriguez_2020_WACV}} &  $51.28$ & $33.04$ & $19.26$ & $-$ & $-$ & $-$ \\
			\multicolumn{2}{|c|}{VSLNet~\cite{zhang2020span}} & $\mathbf{63.16}$ & $\mathit{43.22}$ & $\mathit{26.16}$ & $-$ & $-$ & $-$ \\
			\hline
			\multicolumn{2}{|c|}{\textbf{MS-2D-TAN}} & $\REVISION{62.09}$ & $\REVISION{\mathbf{45.50}}$ & $\REVISION{\mathbf{28.28}}$ & $\REVISION{\mathbf{87.61}}$ & $\REVISION{\mathbf{79.36}}$ & $\REVISION{\mathbf{61.70}}$ \\
			\hline
		\end{tabular}
	\end{center}
	\label{tab:activitynet}
\end{table}

\begin{table}[t]
	\caption{Performance comparison on TACoS.
	}
\setlength{\tabcolsep}{1.0pt}
	\begin{center}
		\begin{tabular}{|c|c|c|c|c|c|c|c|c|c|}
			\hline
			\multicolumn{2}{|c|}{\multirow{2}*{Method}} & \multicolumn{4}{c|}{$Rank1@$} & \multicolumn{4}{c|}{$Rank5@$} \\
			\cline{3-10}
			\multicolumn{2}{|c|}{} & $0.1$ & $0.3$ & $0.5$ & $0.7$ & $0.1$ & $0.3$ & $0.5$ & $0.7$ \\
			\hline
			\multicolumn{10}{|c|}{VGG features} \\
			\hline
			\multicolumn{2}{|@{}c@{}|}{MCN~\cite{hendricks17iccv}} & $14.42$ & $-$ & $5.58$ & $-$ & $37.35$ & $-$ & $10.33$ & $-$  \\
			\multicolumn{2}{|c|}{SM-RL~\cite{wang2019language}} & $26.51$ & $\mathit{20.25}$ & $15.95$ & $-$ & $50.01$ & $38.47$ & $27.84$  & $-$ \\
			\multicolumn{2}{|c|}{SAP~\cite{chen2019semantic}} & $\mathit{31.15}$ & $-$ & $\mathit{18.24}$ & $-$ & $\mathit{53.51}$ & $-$ & $\mathit{28.11}$ & $-$  \\
			\hline
			\multicolumn{2}{|c|}{\textbf{MS-2D-TAN}} & $\REVISION{\mathbf{50.64}}$ & $\REVISION{\mathbf{43.31}}$ & $\REVISION{\mathbf{35.27}}$ & $\REVISION{\mathbf{23.54}}$ & $\REVISION{\mathbf{78.31}}$ &  $\REVISION{\mathbf{66.18}}$ &  $\REVISION{\mathbf{55.81}}$ &  $\REVISION{\mathbf{38.09}}$ \\
% 			\multicolumn{2}{|c|}{\textbf{MS-2D-TAN}*} & $\mathbf{52.89}$ & $\mathbf{44.76}$ & $\mathbf{35.12}$ & $\mathbf{22.77}$ & $\mathbf{79.63}$ &  $\mathbf{68.38}$ &  $\mathbf{55.61}$ &  $\mathbf{35.77}$ \\
			\hline
			\hline
			\multicolumn{10}{|c|}{C3D features} \\
			\hline
			\multicolumn{2}{|c|}{CTRL~\cite{gao2017tall}} & $24.32$ & $18.32$ & $13.30$ & $6.96$ & $48.73$ & $36.69$ & $25.42$ & $15.33$ \\
			\multicolumn{2}{|c|}{MCF~\cite{wu2018multi}} & $25.84$ & $18.64$ & $12.53$ & $-$ & $52.96$ & $37.13$ & $24.73$ & $-$ \\
			\multicolumn{2}{|c|}{TGN~\cite{chen2018temporally}} & $\mathit{41.87}$ & $21.77$ & $18.9$ & $11.88$ & $53.40$ & $39.06$ & $31.02$ & $15.26$\\
			\multicolumn{2}{|c|}{ACRN~\cite{liu2018attentive}} & $24.22$ & $19.52$ & $14.62$ & $-$ & $47.42$ & $34.97$ & $24.88$ & $-$  \\
			\multicolumn{2}{|c|}{ROLE~\cite{liu2018crossmodal}} & $20.37$ & $15.38$ & $9.94$ & $-$ & $45.45$ & $31.17$ & $20.13$ & $-$ \\
			\multicolumn{2}{|c|}{VAL~\cite{song2018val}} & $25.74$ & $19.76$ & $14.74$ & $-$ & $51.87$ & $38.55$ & $26.52$  & $-$ \\
% 			\multicolumn{2}{|c|}{BSSTL~\cite{li2019bidirectional}} & $43.11$ & $22.31$ & $18.73$ & $-$ & $54.67$ & $40.87$ & $29.89$  & $-$ \\
			\multicolumn{2}{|c|}{ACL-K~\cite{Ge_2019_WACV}} & $31.64$ & $24.17$ & $20.01$ & $-$ &  $57.85$ & $42.15$ & $30.66$  & $-$ \\
			\multicolumn{2}{|c|}{DEBUG~\cite{lu2019debug}} & $41.15$ & $23.45$ & $-$ & $-$ & $-$ & $-$ & $-$ & $-$ \\
			\multicolumn{2}{|c|}{GDP~\cite{chen2020rethinking}} & $39.68$ & $24.14$ & $-$ & $-$ & $-$ & $-$ & $-$ & $-$ \\
			\multicolumn{2}{|c|}{CMIN~\cite{zhang2019cross}} & $32.48$ & $24.64$ & $18.05$ & $-$ & $\mathit{62.13}$ & $38.46$ & $27.02$ & $-$ \\
			\multicolumn{2}{|c|}{QSPN~\cite{xu2019multilevel}} & $25.31$ & $20.15$ & $15.23$ & $-$ & $53.21$ & $36.72$ & $25.30$ & $-$ \\
			\multicolumn{2}{|c|}{SLTA~\cite{jiang2019cross}} & $23.13$ & $17.07$ & $11.92$ & $-$ & $46.52$ & $32.90$ & $20.86$  & $-$ \\
			\multicolumn{2}{|c|}{ABLR~\cite{yuan2019to}} & $34.70$ & $19.50$ & $9.40$ & $-$ & $-$ & $-$ & $-$ & $-$ \\
			\multicolumn{2}{|c|}{TripNet~\cite{Hahn2019tripping}} & $-$ & $23.95$ & $19.17$ & $9.52$ & $-$ & $-$ & $-$ & $-$ \\
			\multicolumn{2}{|c|}{SCDM~\cite{yuan2019semantic}} & $-$ & $26.11$ & $21.17$ & $-$ & $-$ & $40.16$ & $32.18$ & $-$ \\
			\multicolumn{2}{|c|}{CBP~\cite{wang2020temporally}} & $-$ & $\mathit{27.31}$ & $\mathit{24.79}$ & $\mathit{19.10}$ & $-$ & $\mathit{43.64}$ & $\mathit{37.40}$ & $\mathit{25.59}$\\
		    \multicolumn{2}{|c|}{DRN~\cite{zeng2020dense}} & $-$ & $-$ & $23.17$ & $-$ & $-$ & $-$ & $33.36$ & $-$ \\
		  %  \multicolumn{2}{|c|}{DPIN~\cite{wang2020dual}} & $59.04$ & $46.74$ & $32.92$ & $-$ & $75.78$ & $62.16$ & $50.26$ & $-$ \\
			\hline
% 			\multicolumn{2}{|c|}{\textbf{MS-2D-TAN}} & $\textbf{49.24}$ & $\textbf{41.74}$ & $\textbf{34.29}$ & $\textbf{21.54}$ & $\textbf{78.33}$ &  $\textbf{67.01}$ &  $\textbf{56.76}$ &  $\textbf{36.84}$ \\
			\multicolumn{2}{|c|}{\textbf{MS-2D-TAN}} & $\REVISION{49.24}$ & $\REVISION{41.74}$ & $\REVISION{34.29}$ & $\REVISION{21.54}$ & $\REVISION{78.33}$ &  $\REVISION{67.01}$ &  $\REVISION{56.76}$ &  $\REVISION{\mathbf{36.84}}$ \\
			\multicolumn{2}{|c|}{\textbf{MS-2D-TAN}$^\star$} & $\REVISION{\mathbf{52.39}}$ & $\REVISION{\mathbf{45.61}}$ & $\REVISION{\mathbf{35.77}}$ & $\REVISION{\mathbf{23.44}}$ & $\REVISION{\mathbf{79.26}}$ &  $\REVISION{\mathbf{69.11}}$ &  $\REVISION{\textbf{57.31}}$ &  $\REVISION{36.09}$ \\
			\hline
			\hline
			\multicolumn{10}{|c|}{I3D features} \\
			\hline
			\multicolumn{2}{|c|}{ExCL~\cite{ghosh2019excl}} & $-$ & $\mathit{45.50}$ & $\mathit{28.00}$ & $13.80$ & $-$ & $-$ & $-$ & $-$ \\
			\multicolumn{2}{|c|}{TMLGA~\cite{Rodriguez_2020_WACV}} & $-$ & $24.54$ & $21.65$ & $16.46$ & $-$ & $-$ & $-$ & $-$ \\
			\multicolumn{2}{|c|}{VSLNet~\cite{zhang2020span}} & $-$ & $29.61$ & $24.27$ & $\mathit{20.03}$ & $-$ &  $-$ &  $-$ &  $-$ \\
			\hline
			\multicolumn{2}{|c|}{\textbf{MS-2D-TAN}} & $\REVISION{48.66}$ & $\REVISION{41.96}$ & $\REVISION{33.59}$ & $\REVISION{22.14}$ & $\REVISION{75.96}$ &  $\REVISION{64.93}$ &  $\REVISION{53.44}$ & $\REVISION{36.12}$ \\
% 			\multicolumn{2}{|c|}{\textbf{MS-2D-TAN}} & $\mathbf{48.66}$ & $\mathit{41.96}$ & $\mathbf{33.59}$ & $\mathbf{22.14}$ & $\mathbf{75.96}$ &  $\mathbf{64.93}$ &  $\mathbf{53.44}$ & $\mathbf{36.12}$ \\
			\multicolumn{2}{|c|}{\textbf{MS-2D-TAN}$^\star$} & $\REVISION{\mathbf{53.24}}$ & $\REVISION{\mathbf{45.96}}$ & $\REVISION{\mathbf{36.59}}$ & $\REVISION{\mathbf{24.79}}$ & $\REVISION{\mathbf{78.73}}$ &  $\REVISION{\mathbf{68.53}}$ &  $\REVISION{\mathbf{57.99}}$ & $\REVISION{\mathbf{37.94}}$ \\
			\hline
		\end{tabular}
	\end{center}
	\label{tab:tacos}
\end{table}

We evaluate the proposed MS-2D-TAN approach on three benchmark datasets, and compare it with recently proposed state-of-the-art methods~\footnote{In addition to the video features, SLTA~\cite{jiang2019cross}, SM-RL~\cite{wang2019language} use extra object features extracted from Faster R-CNN~\cite{ren2015faster}.
}, including:
% \begin{itemize}[leftmargin=0.38cm]
% \item{\textbf{Proposal based methods}:
% \begin{itemize}
%     \item{context concatenation: MCN~\cite{hendricks17iccv},  CTRL~\cite{gao2017tall}, ACRN~\cite{liu2018attentive},  ACL-K~\cite{Ge_2019_WACV},  VAL~\cite{song2018val}}
%     \item{sequential modeling: TGN~\cite{chen2018temporally}, CMIN~\cite{zhang2019cross}, CBP~\cite{wang2020temporally}}
%     \item{graphical modeling: MAN~\cite{zhang2019man}}
%     \item{others: SCDM~\cite{yuan2019semantic},}
% \end{itemize}
% }
% \item{\textbf{Proposal free methods}: ABLR~\cite{yuan2019to}, ExCL~\cite{ghosh2019excl}
% }
% \item{\textbf{Reinforcement learning based methods}: RWM-RL~\cite{he2019read}, SM-RL~\cite{wang2019language}, TripNet~\cite{Hahn2019tripping}, TSP-RPL~\cite{wu2020tree}
% }
% \end{itemize}

\begin{itemize}[leftmargin=0.38cm]
\item {\textbf{clip-based methods}:
    \begin{itemize}[leftmargin=0.38cm]
    \item{\textbf{anchor based methods}: 
    TGN~\cite{chen2018temporally}, CMIN~\cite{zhang2019cross} and CBP~\cite{wang2020temporally}, SCDM~\cite{yuan2019semantic}},
    \item {\textbf{anchor free methods}:
    ACRN~\cite{liu2018attentive}, ROLE~\cite{liu2018crossmodal}, SLTA~\cite{jiang2019cross}, DEBUG~\cite{lu2019debug}, VSLNet~\cite{zhang2020span}, GDP~\cite{chen2020rethinking}, LGI~\cite{mun2020local}, ABLR~\cite{yuan2019to}, TMLGA~\cite{Rodriguez_2020_WACV}, ExCL~\cite{ghosh2019excl} and DRN~\cite{zeng2020dense}},
    \item {\textbf{reinforcement learning based methods}:
    RWM-RL~\cite{he2019read}, SM-RL~\cite{wang2019language}, TripNet~\cite{Hahn2019tripping} and TSP-RPL~\cite{wu2020tree}},
    \end{itemize}
}

\item {\textbf{moment-based methods}:
    \begin{itemize}[leftmargin=0.38cm]
    \item{\textbf{global/local context methods}: 
    MCN~\cite{hendricks17iccv},  CTRL~\cite{gao2017tall},  ACL-K~\cite{Ge_2019_WACV} and  VAL~\cite{song2018val}
    },
    \item{\textbf{graph based methods}: MAN~\cite{zhang2019man}},
    \item{\textbf{cascade refinement methods}: QSPN~\cite{xu2019multilevel} and SAP~\cite{chen2019semantic}}.
    \end{itemize}
}
\end{itemize}
The results are summarized in Table~\ref{tab:charades}--~\ref{tab:tacos}. The values highlighted
by \textbf{bold} and \textit{italic} fonts indicate the top-2 methods, respectively. All results are reported in percentage $(\%)$.

The results show that MS-2D-TAN performs the best in various scenarios on all three benchmark datasets across different criteria. In most cases, MS-2D-TAN ranks the first or the second place~\footnote{ Comparing CBP on $Rank5@0.7$ in Table~\ref{tab:charades}, our MS-2D-TAN is worse due to the the IoU threshold for NMS. If the threshold is to set $0.5$, the $Rank5@0.5$ and $Rank5@0.7$ of our model are $77.58$ and $53.04$, which outperform CBP.}. It is worth noting that on TACoS dataset (see Table~\ref{tab:tacos}), our MS-2D-TAN surpasses the previous best approach CBP~\cite{wang2020temporally}
, by approximate $18$ points and $25$ points in term of $Rank1@0.3$ and $Rank5@0.3$, respectively. Moreover, on the large-scale ActivityNet Captions dataset, MS-2D-TAN also outperforms the top ranked method DRN~\cite{zeng2020dense} and VSLNet~\cite{zhang2020span} with respect to $IoU@0.5$ and $0.7$. It validates that MS-2D-TAN is able to localize the moment boundary more precisely.

In more details, by comparing MS-2D-TAN with other related methods, we obtain several observations.
First, we compare MS-2D-TAN with previous moment-based methods: MCN~\cite{hendricks17iccv},  CTRL~\cite{gao2017tall}, ACL-K~\cite{Ge_2019_WACV},  VAL~\cite{song2018val} and MAN~\cite{zhang2019man}.
From the results in Table~\ref{tab:charades}--\ref{tab:tacos}, 
we observe that our MS-2D-TAN achieves superior results than concatenating local/global context. The reason is that independently matching the sentence with moment candidates ignores the temporal contexts, and cannot distinguish the small differences between the overlapped moments.
Differently, our proposed MS-2D-TAN models the relations between moment candidates by a series of sparse 2D temporal maps, and enables the network to perceive more context information from the adjacent moment candidates. Hence, it gains large improvements compared to the  methods that only consider global or local context.
The closest work, MAN~\cite{zhang2019man}, utilizes GCN to model the relations among the moments and outperforms its previous methods on Charades-STA with VGG features. 
The experiment results demonstrate the effectiveness of the moment-level relation modeling.
However, the nodes in GCN are permutation invariant, thus they ignore the temporal ordering of different moments.
In contrast, our MS-2D-TAN models the temporal relations through a 2D convolution network and it can naturally learn the temporal ordering through its convolution kernel.
From Table~\ref{tab:charades}, we can observe that MS-2D-TAN surpasses MAN in all evaluation metrics.

Moreover, we compare our approach with clip-based methods (including anchor-based methods, anchor-free methods and RL-based methods).
Previous anchor-based methods include TGN~\cite{chen2018temporally}, CMIN~\cite{zhang2019cross}, CBP~\cite{wang2020temporally} and SCDM~\cite{yuan2019semantic}.
Due to the involvement of more context information during prediction, the anchor-based approaches perform better than the global/local context approaches, however, inferior to our proposed MS-2D-TAN method. 
Anchor-based approaches implicitly learn the moment context information through a recurrent memory module or a 1D convolution network, while our MS-2D-TAN explicitly exploits the long range context information via the sparse multi-scale 2D temporal map. 
It further verifies the effectiveness of our model in high quality moment localization.
We also compare our method with the anchor-free methods~\cite{liu2018attentive,liu2018crossmodal,jiang2019cross,lu2019debug,zhang2020span,chen2020rethinking,mun2020local,yuan2019to,Rodriguez_2020_WACV,ghosh2019excl,zeng2020dense}, reinforcement learning based methods~\cite{he2019read,wang2019language,Hahn2019tripping,wu2020tree} and cascade refinement methods~\cite{xu2019multilevel,chen2019semantic}.
Noted that anchor free methods and reinforcement learning methods do not rely on predefined anchors and directly predict the start and end time stamp. Therefore, their upper bounds under any IoU values should all be $100\%$, which is not directly comparable with our methods. 
Even with lower upper bound, our model still achieves superior performance compared to theirs.
Compared with cascade refinement methods, our method predicts moments in one stage. While it is simple in design, it still outperforms these two methods.
This demonstrates the significance of introducing moment context information.
We also find that the performance improvement in TACoS dataset is larger than the other two datasets.
This also verifies that our MS-2D-TAN is more efficient in modeling long videos.

% We also find that Charades-STA is more sensitive to the visual feature choices compared to the other two datasets.

\subsection{Ablation Study}

We conduct all our experiments with C3D features in our ablation study.
% We have shown the best hyper parameter settings of all three datasets in Table~\ref{tab:setting}.
In this section, we first verify the efficiency of our MS-2D-TAN from both theoretical proof and experiments.
We then verify the effectiveness of the sparse multi-scale map and gated convolution of our model in Table~\ref{tab:ablation_study} \footnote{We use the finetuned \REVISION{hyperparameters} ($H=512$, $N=512$, $A=8$, $\kappa=9$, and $L=2$) for the ablation study on TACoS in Table~\ref{tab:ablation_study} and Table~\ref{tab:tacos_simple_complex}, while others follow the default setting ($H=512$, $N=64$, $A=16$, $\kappa=17$, and $L=2$).}. 
\REVISION{We also evaluate how does our MS-2D-TAN perform under different types of queries in Table~\ref{tab:tacos_simple_complex}, and compare the sparse multi-scale map and the dense single-scale map under the same receptive field in Table~\ref{tab:receptive_field}.
Finally, we conduct ablation studies on the \REVISION{hyperparameters} of our model in Table~\ref{tab:hyperparameters}, \textit{i.e.}, the kernel and layer settings, the number of anchors, window size and the number of hidden states.}
We also compare the number of candidate moments with previous methods.
% We believe that using different combination may further improve the performance.

\REVISION{In Table~\ref{tab:ablation_study} - \ref{tab:hyperparameters}, Pool and Conv in the ``Feat'' column indicate whether using max pooling or stacked convolutions for moment feature extraction. 
DS, SS and MS in the ``Map'' column represent the dense single-scale map, sparse single-scale map and the sparse multi-scale map. 
C, G and \xmark~in the ``TAN'' column represent 2D-TAN with convolution layers, 2D-TAN with gated convolution layers and without using 2D-TAN.
$H$, $N$, and \REVISION{$K$} represent the size of each hidden layer, sliding window size, and the number of scales. $A$, \REVISION{$\kappa$}, and $L$ represent the number of anchors, kernel size, and the number of layers of gated convolutions at each scale, respectively.
}

\begin{table*}[t]
	\caption{Ablation study on the feature extraction method, the type of 2D temporal map and 2D-TAN. 
% 	Pool and Conv in the ``Feat'' column indicate whether using max pooling or stacked convolutions for moment feature extraction. DS, SS and MS in the ``Map'' column represent the dense single-scale map, sparse single-scale map and the sparse multi-scale map. C and G in the ``TAN'' column represent 2D-TAN with convolution layers and gated convolution layers.
	}
	\setlength{\tabcolsep}{.9pt}
	\begin{center}
		\begin{tabular}{|c|c|c|c|c|c|c|c|c|c|c|c|c|c|c|c|c|c|c|c|c|c|}
			\hline
			\multirow{3}*{Row\#} & \multirow{3}*{Feat} & \multirow{3}*{Map} & \multirow{3}*{TAN} &  \multicolumn{4}{c|}{Charades-STA} & \multicolumn{6}{c|}{ActivityNet Captions} & \multicolumn{8}{c|}{TACoS}\\
            \cline{5-22}
			& & & & \multicolumn{2}{c|}{$Rank1@$} & \multicolumn{2}{c|}{$Rank5@$} & \multicolumn{3}{c|}{$Rank1@$} &
			\multicolumn{3}{c|}{$Rank5@$} &
			\multicolumn{4}{c|}{$Rank1@$} &
			\multicolumn{4}{c|}{$Rank5@$} \\
			\cline{5-22}
			& & & & $0.5$ & $0.7$ & $0.5$ & $0.7$ & $0.3$ & $0.5$ & $0.7$ & $0.3$ & $0.5$ & $0.7$ & $0.1$ & $0.3$ & $0.5$ & $0.7$ & $0.1$ & $0.3$ & $0.5$ & $0.7$\\
			\hline
			$1$ & Conv & DS & C & $39.49$ & $20.83$ & $80.59$ & ${50.11}$ & $\REVISION{60.29}$ & $\mathit{\REVISION{45.82}}$ & $\mathit{\REVISION{29.08}}$ & $\REVISION{85.94}$ & $\REVISION{77.08}$ & $\mathbf{\REVISION{61.39}}$ & $41.74$ & $32.74$ & $23.09$ & $14.10$ & $73.08$ & $56.89$ & $43.36$ & $25.04$ \\
			$2$ & Conv & SS & C & $\REVISION{39.95}$ & $\REVISION{21.83}$ & $\REVISION{80.97}$ & $\mathit{\REVISION{54.09}}$ & $\REVISION{60.10}$ & $\REVISION{44.45}$ & $\REVISION{26.94}$ & $\REVISION{86.84}$ & $\REVISION{78.36}$ & $\REVISION{59.44}$ & $\REVISION{44.89}$ & $\REVISION{37.19}$ & $\REVISION{28.57}$ & $\REVISION{18.30}$ & $\REVISION{73.11}$ & $\REVISION{62.26}$ & $\REVISION{49.86}$ & $\REVISION{30.84}$ \\
			$3$ & Conv & MS & C & ${40.97}$ & ${22.55}$ & ${80.91}$ & $46.48$ & $\mathit{\REVISION{60.75}}$ & $\REVISION{45.37}$ & $\REVISION{28.25}$ & $\REVISION{87.05}$ & $\mathbf{\REVISION{78.89}}$ & $\REVISION{60.67}$ & ${51.64}$ & ${44.61}$ & ${35.29}$ & ${23.09}$ & $\mathit{79.23}$ & $\mathit{68.63}$ & ${56.04}$ & ${35.57}$ \\
			$4$ & Conv & MS & G & $\mathit{41.10}$ & $\mathit{23.25}$ & $\mathit{81.53}$ & ${48.55}$ & $\mathbf{\REVISION{61.04}}$ & $\mathbf{\REVISION{46.16}}$ & $\mathbf{\REVISION{29.21}}$ & $\mathit{\REVISION{87.30}}$ & $\mathit{\REVISION{78.80}}$ & $\mathit{\REVISION{60.85}}$ & $\mathit{52.39}$ & $\mathbf{45.61}$ & $\mathit{35.77}$ & $\mathit{23.44}$ & $\mathbf{79.26}$ &  $\mathbf{69.11}$ &  $\mathit{57.31}$ &  $\mathit{36.09}$ \\	
			$5$ & Pool & MS & G & $\mathbf{\REVISION{42.42}}$ & $\mathbf{\REVISION{25.11}}$ & $\mathbf{\REVISION{84.38}}$ & $\mathbf{\REVISION{54.49}}$ & $\REVISION{60.51}$ & $\REVISION{43.59}$ & $\REVISION{25.58}$ & $\mathbf{\REVISION{87.51}}$ & $\REVISION{78.72}$ & $\REVISION{60.16}$ & $\mathbf{\REVISION{52.84}}$ & $\mathit{\REVISION{45.34}}$ & $\mathbf{\REVISION{36.34}}$ & $\mathbf{\REVISION{24.47}}$ & $\REVISION{78.58}$ & $\REVISION{68.06}$ & $\mathbf{\REVISION{57.56}}$ & $\mathbf{\REVISION{37.54}}$ \\
% 			$6$ & Conv & MS & S & $\mathbf{42.88}$ & $\mathit{24.97}$ & $\mathit{81.56}$ & ${49.54}$ & $60.57$ & $46.11$ & $\mathit{29.40}$ & $86.42$ & $77.17$ & $59.42$ & $\mathit{52.64}$ & $44.34$ & $\mathit{35.92}$ & $\mathbf{24.64}$ & $78.36$ &  $67.83$ &  $56.86$ & $\mathbf{38.39}$ \\		
			\hline
		\end{tabular}
	\end{center}

	\label{tab:ablation_study}
\end{table*}

\subsubsection{Memory Usage and Speed}
\label{sec:memory_speed}
\begin{figure}[t]
    \centering
    \includegraphics[width=\columnwidth]{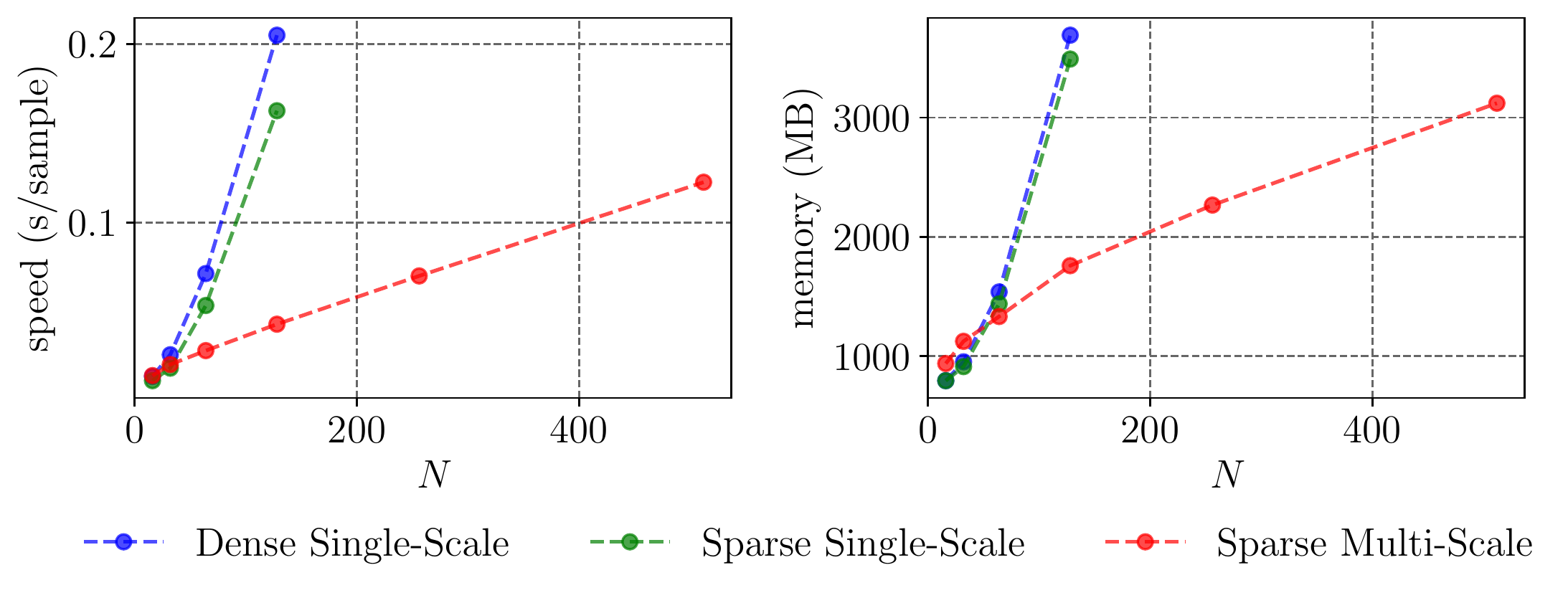}
    \caption{\REVISION{Comparisons on inference speed and memory cost of three types of feature map. $N$ is the number of video clips. \textbf{Left:} inference speed. \textbf{Right:} memory cost.}}
    \label{fig:speed_memory_self}
\end{figure}

\begin{figure}[t]
    \centering
    \includegraphics[width=\columnwidth]{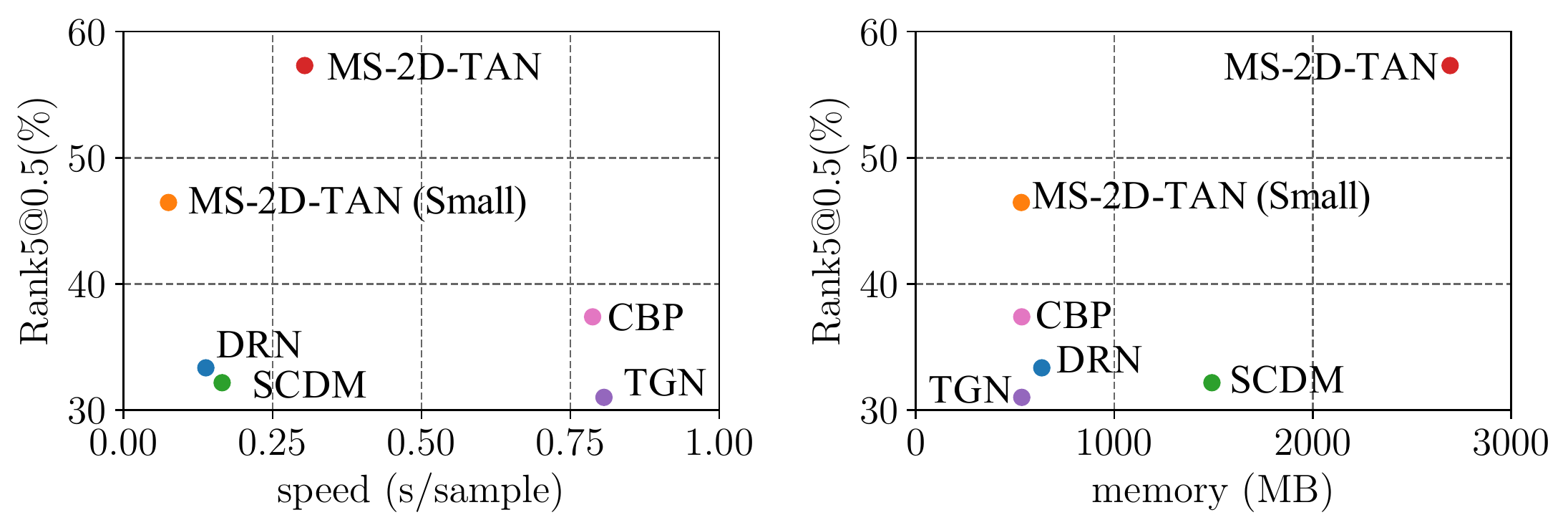}
    \caption{\REVISION{Comparisons on inference speed and memory cost with other baselines on TACoS. \textbf{Left:} inference speed. \textbf{Right:} memory cost. MS-2D-TAN (small) is a MS-2D-TAN model with small size, where its hidden units are set to $H=64$.}}
    \label{fig:speed_memory_others}
\end{figure}

In this section, we compare the speed and memory cost of the dense single-scale map, the sparse single-scale map, the sparse multi-scale map, as well as other baselines. 
In the moment feature extraction module, the total number of moment candidates for dense single-scale map is $\sum_{i=1}^Ni=O(N^2)$.
Meanwhile, the total number of moment candidates for sparse single-scale map is 
\begin{equation}
    \begin{aligned}
    &A\cdot N + A\cdot \frac{N}{2} + ... + A\cdot \frac{N}{2^{K-1}} \\
    =& (2-\frac{1}{2^{K-1}}) AN\\
    =& O(N),\\
    \end{aligned}
\end{equation}
where the number of video clips $N$ is much larger than the number of anchors $A$ and the number of scales $K$ in most cases. 
Therefore, we can reduce the computational cost of the moment feature extraction module from $O(N^2)$ to $O(N)$ by using sparse sampling.

\REVISION{
However, only reducing the computational cost of moment feature extraction is not enough. In our preliminary work \cite{2DTAN_2020_AAAI}, we reconstruct a sparse single-scale map with a size of $N^2$ and feed it to 2D-TAN. Therefore, 2D-TAN still requires time and memory complexities of $O(N^2)$.
Our proposed multi-scale extension can tackle this problem by separating it into multiple feature maps of size $A\cdot N$, $A\cdot \frac{N}{2}$, ..., $A\cdot \frac{N}{2^{K-1}}$, thus reducing the time and memory complexities of 2D-TAN from $O(N^2)$ to $O(N)$~\footnote{Since both the dilation and stride are equal across different scales in our implementation, we reformat the $k$-th sparse maps ($k>0$) to dense maps with a stride of $2^k$ and do convolution operation with a stride of $1$ and a dilation of $1$. This is equal to the previous, however, using less memory cost.}.
}

\REVISION{
We measure the computational cost of the three types of maps through experiments~\footnote{Since the CuDNN package optimizes memory usage at run time, we turn it off for this comparison.}.
As shown in Fig.~\ref{fig:speed_memory_self}, the time and memory usage of the dense single-scale map increases quadratically as the sequences become longer. 
The sparse single-scale map is better than dense single-scale map, however, its time and memory usage still increase quadratically.
In contrast, the time and the memory usage of the sparse multi-scale map scales linearly with the sequence length.
From both the theoretical and experimental results, we find that the sparse multi-scale modeling has faster speed and lower memory cost compared to the dense single-scale modeling and sparse single-scale modeling.}

\REVISION{
We also compare the speed and memory cost of MS-2D-TAN with several recent approaches, including SCDM~\cite{yuan2019semantic}, DRN~\cite{zeng2020dense}, TGN~\cite{chen2018temporally} and CBP~\cite{wang2020temporally}. From Fig~\ref{fig:speed_memory_others}, we can observe that MS-2D-TAN$^\star$ runs faster with less memory usage and achieves better performance than the aforementioned baselines.
Further increasing the hidden state size of MS-2D-TAN achieves better performance (MS-2D-TAN \textit{v.s.} MS-2D-TAN (Small)) but sacrifices speed and memory usage.
}

\subsubsection{Effectiveness of Sparse Multi-Scale Map}
In this section, we verify the effectiveness of our sparse multi-scale map.
\REVISION{As shown in Table~\ref{tab:ablation_study}, comparing with the dense single-scale map (row~$1$), the sparse single-scale map (row~$2$) achieves competitive results using less number of moment candidates.~\footnote{\SECONDRREVISION{In the three datasets (see Fig. $3$), most of the target moments are short in time. Our sparse sampling strategy retains short moment candidates while discarding most long candidates, which leverages the distribution prior of target moments. Therefore, it obtains better performance in most cases.
(row~$1$ \textit{v.s.} row~$2$)}}
The sparse multi-scale map (row~$4$) further reduces the time and memory cost of the sparse single-scale map, while 
achieving better performance on TACoS and competitive results on Charades-STA and ActivityNet Captions.}
The reason is that the videos in TACoS have longer average duration than the other datasets, as shown in Fig.~\ref{fig:distribution}. 
The dense single-scale map \REVISION{and the sparse single-scale map} cannot handle long videos well, since the receptive field is limited and context information from long range is missed.
In contrast, the sparse multi-scale map has larger receptive field and involves more long range context and therefore outperforms the previous in TACoS dataset.

\subsubsection{Effectiveness of Gated Convolution}
In this section, we verify the effectiveness of the gated convolution in Table~\ref{tab:ablation_study}.
Comparing row~$3$ and row~$4$, we find applying an additional gate on the convolution can improve the performance on TACoS dataset.
The better performance of gated convolution is also consistent with its effectiveness in image inpainting~\cite{yu2019free}.
The gated convolution is more flexible to adjust the weights of moments among different scales compared to the conventional convolution.

\begin{table*}[t!]
    \centering
	\setlength{\tabcolsep}{1.pt}
    \caption{\REVISION{Performance comparison on the type of query on TACoS. 
    }}
	\begin{tabular}{|c|c|c|c|c|c|c|c|c|c|c|c|c|c|c|c|c|c|c|c|}
	\hline
	\multirow{3}*{Row} & \multirow{3}*{Feat} & \multirow{3}*{Map} & \multirow{3}*{TAN} & \multicolumn{8}{c|}{Simple Queries} & \multicolumn{8}{c|}{Complex Queries} \\
    \cline{5-20}
	& & & & \multicolumn{4}{c|}{$Rank1@$} & \multicolumn{4}{c|}{$Rank5@$} & \multicolumn{4}{c|}{$Rank1@$} &
	\multicolumn{4}{c|}{$Rank5@$} \\
	\cline{5-20}
	& & & & $0.1$ & $0.3$ & $0.5$ & $0.7$ & $0.1$ & $0.3$ & $0.5$ & $0.7$ & $0.1$ & $0.3$ & $0.5$ & $0.7$ & $0.1$ & $0.3$ & $0.5$ & $0.7$ \\
	\hline
	$1$ & Conv & DS & \xmark  & $29.53$ & $22.49$ & $15.09$ & $5.41$ & $63.81$ & $51.20$ & $35.77$ & $13.75$ & $28.92$ & $16.06$ & $8.43$ & $2.81$ & $68.67$ & $49.80$ & $28.92$ & $8.84$ \\
	$2$ & Conv & DS & C & $41.95$ & $33.40$ & $23.69$ & $14.74$ & $73.40$ & $57.28$ & $43.66$ & $25.56$ & $38.55$ & $23.29$ & $13.25$ & $4.82$ & $68.67$ & $52.21$ & $39.36$ & $19.28$ \\
	$3$ & Conv & SS & C & $45.52$  & $38.09$  & $29.37$  & $19.00$  & $73.32$  & $62.69$  & $50.67$  & $31.45$  & $36.95$  & $24.90$  & $16.87$  &  $7.63 $ & $71.49$  & $55.82$  & ${41.37}$  & $19.68$ \\
	$4$ & Conv & MS & C & ${52.16}$ & ${45.52}$ & ${36.17}$ & $\mathit{23.91}$ & $\mathit{79.72}$ & $\mathit{69.43}$ & ${57.14}$ & ${36.51}$ & $\mathbf{42.57}$ & ${30.12}$ & $\mathit{22.09}$ & $\mathit{10.04}$ & $\mathit{73.49}$ & $\mathit{57.43}$ & ${40.16}$ & $\mathit{20.08}$ \\
	$5$ & Conv & MS & G  & $\mathit{53.30}$ & $\mathbf{46.46}$ & $\mathit{36.73}$ & $\mathit{24.15}$ & $\mathbf{79.98}$ & $\mathbf{69.96}$ & $\mathit{58.18}$ & $\mathit{36.91}$ & $\mathit{38.15}$ & $\mathbf{32.13}$ & ${21.69}$ & $\mathit{11.24}$ & ${72.29}$ & $\mathit{57.43}$ & $\mathit{44.98}$ & $\mathit{24.10}$ \\
	$6$ & Pool & MS & G & $\mathbf{53.65}$ & $\mathit{46.43}$ & $\mathbf{37.26}$ & $\mathbf{25.03}$ & $78.89$ & $68.60$ & $\mathbf{58.50}$ & $\mathbf{38.14}$ & $\mathit{40.96}$ & $\mathit{30.92}$ & $\mathbf{22.49}$ & $\mathbf{16.06}$ & $\mathbf{75.10}$ & $\mathbf{60.64}$ & $\mathbf{45.78}$ & $\mathbf{29.32}$ \\
	\hline
    \end{tabular}
    \label{tab:tacos_simple_complex}
\end{table*}

\begin{table*}[t!]
	\caption{\REVISION{Ablation study on receptive field size. 
	}}
\setlength{\tabcolsep}{2pt}
	\begin{center}
		\begin{tabular}{|c|c|c|c|c|c|c|c|c|c|c|c|c|c|c|c|c|c|c|c|}
			\hline
			\multirow{2}*{Row\#} & Receptive & \multirow{2}*{Feat}  & \multirow{2}*{Map} & \multirow{2}*{TAN} & \multicolumn{6}{c|}{Hyperparameters} &  \multicolumn{4}{c|}{$Rank1@$} & \multicolumn{4}{c|}{$Rank5@$} \\
			\cline{6-19}
			& Field Size & & & & $H$ & $N$ & $K$ & $A$ & $\kappa$ & $L$ & $0.1$ & $0.3$ & $0.5$ & $0.7$ & $0.1$ & $0.3$ & $0.5$ & $0.7$ \\
			\hline

			$1$ & \multirow{2}*{$17\times 17$} & \multirow{2}*{Conv} & DS & \multirow{2}*{C} & \multirow{2}*{$512$} & \multirow{2}*{$512$} & $1$ & $128$ & $9$ & \multirow{2}*{$2$} & $41.74$ & $32.74$ & $23.09$ & $14.10$ & $\mathit{73.08}$ & $56.89$ & $43.36$ & $25.04$ \\%53415939
			$2$ & & & MS & & & & $2$ & $64$ & $5$ & & $38.72$ & $30.97$ & $21.79$ & $13.15$ & $70.48$ & $56.31$ & $42.64$ & $25.67$ \\%53415939
			\hline
			$3$ & \multirow{2}*{$33\times 33$} & \multirow{2}*{Conv} & DS & \multirow{2}*{C} & \multirow{2}*{$512$} & \multirow{2}*{$512$} & $1$ & $128$ & $17$ & \multirow{2}*{$2$} & $43.11$ & $34.49$ & $25.79$ & $16.65$ & $70.86$ & $58.21$ & $45.96$ & $\mathit{29.47}$ \\
			$4$ & & & MS & & & & $3$ & $32$ & $5$ & & $38.84$ & $31.67$ & $22.27$ & $14.72$ & $72.86$ & $58.89$ & $43.84$ & $25.69$ \\%53415939
			\hline
			$5$ & \multirow{2}*{$65\times 65$} & \multirow{2}*{Conv} & DS & \multirow{2}*{C} & \multirow{2}*{$512$} & \multirow{2}*{$512$} & $1$ & $128$ & $33$ & \multirow{2}*{$2$} & $-$ & $-$ & $-$ & $-$ & $-$ & $-$ & $-$ & $-$ \\%53415939
			$6$ & & & MS & & & & $4$ & $16$ & $5$ & & $\mathit{44.29}$ & $\mathit{35.64}$ & $\mathit{26.72}$ & $\mathit{16.97}$ & $72.98$ & $\mathit{60.36}$ & $\mathit{47.54}$ & $29.07$ \\%53415939
			\hline
			$7$ & \multirow{2}*{$129\times 129$} & \multirow{2}*{Conv} & DS & \multirow{2}*{C} & \multirow{2}*{$512$} & \multirow{2}*{$512$} & $1$ & $128$ & $65$ & \multirow{2}*{$2$} & $-$ & $-$ & $-$ & $-$ & $-$ & $-$ & $-$ & $-$ \\%53415939
			$8$ & & & MS & & & & $5$ & $8$ & $5$ & & $
			\mathbf{45.89}$ & $\mathbf{39.29}$ & $\mathbf{31.32}$ & $\mathbf{20.24}$ & $\mathbf{73.36}$ & $\mathbf{61.83}$ & $\mathbf{50.66}$ & $\mathbf{32.37}$ \\%53415939
            \hline

		\end{tabular}
	\end{center}
	\label{tab:receptive_field}
\end{table*}

\subsubsection{Max Pooling \textit{v.s.} Stacked Convolution}
\REVISION{
In this section, we compare the usage of max pooling and stacked convolution for moment feature extraction. As shown in Table~\ref{tab:ablation_study}, we observe that using max pooling has similar performance with stacked convolutions on ActivityNet Captions and TACoS, while getting better performance on Charades-STA. The max-pooling operation is fast in calculation, since it does not contain any parameters, thus it is suitable for computation intensive applications.
}

\subsubsection{Performance on Different Types of Query}

\REVISION{
In this section, we investigate how our proposed model contributes to different types of queries.
 Following previous works~\cite{hendricks18emnlp,zhang2019exploiting}, we define complex queries as sentences that include temporal keywords, such as ``before'', ``while'', ``then'', ``after'', ``continue'' and ``again''. All the rest sentences are regarded as simple queries. This rule separates sentences into $3,752$ simple queries and $249$ complex queries on TACoS.
The results are shown in Table~\ref{tab:tacos_simple_complex}. Comparing row $1$-$2$, we can observe that using 2D-TAN benefits both simple queries and complex queries.
Comparing the row $2$-$4$, using our multi-scale extension can further improve the performance, where the largest improvement comes from simple queries.
Comparing the row $4$-$6$, we can observe that the gated convolution can further improve MS-2D-TAN on simple queries and maintain comparable performance on complex queries.
Comparing row $5$-$6$, using max pooling achieves better performance on complex queries and similar performance on simple queries.
These experiments verifies that our multi-scale extension can improve localizing simple queries as well as complex queries with temporal relations. 
Besides, tuning on the feature extraction method and the convolution type of 2D-TAN may obtain better performance on a specific dataset.
}

\subsubsection{Effectiveness of Receptive Field}
\REVISION{
To investigate whether the model improvements comes from the large receptive field or the multi-scale context modeling, we conduct an ablation study in Table~\ref{tab:receptive_field}.
To increase receptive fields, the dense single-scale map uses larger kernel size $\kappa$, while sparse single-scale map uses larger number of scales $K$. As shown in row~$1$-$4$, within the same receptive field, using a dense single-scale map is slightly better than using a sparse multi-scale map. 
However, when the receptive field is large enough (row~$3$ and row~$4$), the dense single-scale map is not feasible since the time and memory cost increase a lot. In contrast, the sparse multi-scale map can further benefit from the larger receptive fields and achieves a better performance.
}

\begin{table*}[t]
	\caption{Ablation study on \REVISION{hyperparameters}. 
% 	$H$, $N$, and \REVISION{$K$} represent the size of each hidden layer, sliding window size, and the number of scales. $A$, \REVISION{$\kappa$}, and $L$ represent the number of anchors, kernel size, and the number of layers of gated convolutions at each scale, respectively.
	}
\setlength{\tabcolsep}{2pt}
	\begin{center}
		\begin{tabular}{|c|c|c|c|c|c|c|c|c|c|c|c|c|c|c|c|c|}
			\hline
			\multirow{2}*{Row\#} & \multirow{2}*{Method} & \multicolumn{6}{c|}{\REVISION{Hyperparameters}}  & \multicolumn{4}{c|}{$Rank1@$} & \multicolumn{4}{c|}{$Rank5@$} & \multirow{2}*{\#Params ($\times10^6$)} \\
			\cline{3-16}
			& & $H$ & $N$ & $K$ & $A$ & $\kappa$ & $L$ & $0.1$ & $0.3$ & $0.5$ & $0.7$ & $0.1$ & $0.3$ & $0.5$ & $0.7$ & \\
			\hline
			$1$ & \multirow{10}*{Upper Bound} & $-$ & $-$ & $1$ & $128$ & $-$ & $-$  & $100.00$ & $99.40$ & $98.95$ & $96.38$ & $100.00$ & $99.40$ & $98.95$ & $96.38$ & $-$ \\
% 			$4$ & Upper Bound & $-$ & $-$ & $2$ & $64$ & $-$ & $-$ & $100.00$ & $99.85$ & $99.43$ & $98.15$ & $100.00$ & $99.85$ & $99.43$ & $98.15$ \\
% 			$5$ & Upper Bound & $-$ & $-$ & $3$ & $32$ & $-$ & $-$ & $100.00$ & $99.85$ & $99.44$ & $98.15$ & $100.00$ & $99.85$ & $99.44$ & $98.15$ \\
			$2$ & & $-$ & $-$ & $4$ & $16$ & $-$ & $-$ & $100.00$ & $99.40$  & $98.95$ & $96.38$ & $100.00$ & $99.40$  & $98.95$ & $96.38$ & $-$ \\
			$3$ & & $-$ & $-$ & $1$ & $8$ & $-$ & $-$ & $91.90$ & $68.53$ & $54.16$ & $41.61$ & $91.90$ & $68.53$ & $54.16$ & $41.61$ & $-$ \\
			$4$ & & $-$ & $-$ & $2$ & $8$ & $-$ & $-$ & $96.68$ & $85.80$ & $72.81$ & $64.31$ & $96.68$ & $85.80$ & $72.81$ & $64.31$ & $-$ \\
			$5$ & & $-$ & $-$ & $3$ & $8$ & $-$ & $-$ & $99.23$ & $94.65$ & $89.35$ & $80.93$ & $99.23$ & $94.65$ & $89.35$ & $80.93$ & $-$ \\
			$6$ & & $-$ & $-$ & $4$ & $8$ & $-$ & $-$ & $100.00$ & $98.10$ & $95.65$ & $92.58$ & $100.00$ & $98.10$ & $95.65$ & $92.58$ & $-$ \\
			$7$ & & $-$ & $-$ & $5$ & $8$ & $-$ & $-$ & $100.00$ & $99.40$ & $98.95$ & $96.38$ & $100.00$ & $99.40$ & $98.95$ & $96.38$ & $-$ \\
			$8$ & & $-$ & $-$ & $6$ & $8$ & $-$ & $-$ & $100.00$ & $100.00$ & $99.75$ & $98.20$ & $100.00$ & $100.00$ & $99.75$ & $98.20$ & $-$ \\
			$9$ & & $-$ & $-$ & $7$ & $8$ & $-$ & $-$ & $100.00$ & $100.00$ & $100.00$ & $98.90$ & $100.00$ & $100.00$ & $100.00$ & $98.90$ & $-$ \\
			$10$ & & $-$ & $-$ & $6$ & $4$ & $-$ & $-$ & $100.00$ & $99.40$ & $98.95$ & $96.35$ & $100.00$ & $99.40$ & $98.95$ & $96.35$ & $-$ \\
% 			$8$ & Upper Bound & $-$ & $-$ & $7$ & $2$ & $-$ & $-$ & $100.00$ & $99.89$ & $99.44$ & $74.00$ & $100.00$ & $99.85$ & $99.44$ & $74.00$ & $-$ \\
			\hline
			$11$ & \multirow{17}*{MS-2D-TAN} & $512$ & $512$ & $4$ & $16$ & $9$ & $2$ & $47.79$ & $39.74$ & $30.64$ & $20.32$ & $76.33$ & $64.53$ & $52.11$ & $33.57$ & $19.87$  \\%198684172
			$12$ & & $512$ & $512$ & $6$ & $4$ & $9$ & $1$ & $45.04$ & $37.24$ & $27.54$ & $16.72$ & $69.26$ & $58.84$ & $48.26$ & $33.84$ & $14.30$  \\%142986764
			$13$ & & $512$ & $512$ & $5$ & $8$ & $1$ & $1$ & $21.04$ & $12.87$ & $6.17$ & $1.87$ & $55.59$ & $39.02$ & $21.94$ & $9.32$ & $21.64$  \\%21644298
			$14$ & & $512$ & $512$ & $5$ & $8$ & $9$ & $1$ & $49.81$ & $41.71$ & $33.07$ & $20.89$ & $76.03$ & $66.63$ & $55.39$ & $36.22$  & $12.67$ \\%126706698
			$15$ & & $512$ & $512$ & $5$ & $8$ & $5$ & $4$ & $49.54$ & $42.59$ & $33.49$ & $23.64$ & $76.03$ & $64.68$ & $53.16$ & $36.64$  & $15.17$ \\%151666713
			$16$ & & $512$ & $512$ & $5$ & $8$ & $17$ & $1$ & $50.61$ & $43.14$ & $35.17$ & $24.32$ & $76.31$ & $66.33$ & $56.54$ & $\mathbf{39.79}$  & $39.99$ \\%399868938
			$17$ & & $512$ & $512$ & $1$ & $8$ & $9$ & $2$ & $40.44$ & $32.52$ & $25.49$ & $17.70$ & $59.39$ & $45.69$ & $35.14$ & $23.84$  & $53.42$ \\%53415939
			$18$ & & $512$ & $512$ & $2$ & $8$ & $9$ & $2$ & $43.14$ & $36.84$ & $30.42$ & $22.44$ & $67.78$ & $55.49$ & $44.26$ & $32.37$ & $98.33$ \\%98333190
			$19$ & & $512$ & $512$ & $3$ & $8$ & $9$ & $2$ & $48.81$ & $42.29$ & $34.09$ & $23.64$ & $73.21$ & $63.01$ & $52.39$ & $34.99$ & \REVISION{$143.25$} \\%143250441
			$20$ & & $512$ & $512$ & $4$ & $8$ & $9$ & $2$ & $50.44$ & $44.04$ & $35.04$ & $23.37$ & $74.33$ & $64.23$ & $53.16$ & $35.44$  & \REVISION{$188.16$} \\%188167692
			$21$ & & $512$ & $512$ & $5$ & $8$ & $9$ & $2$ & $\mathbf{52.39}$ & $\mathbf{45.61}$ & $\mathbf{35.77}$ & $23.44$ & $79.26$ &  $\mathbf{69.11}$ &  $\mathbf{57.31}$ &  $36.09$  & \REVISION{$233.08$} \\%233084943
			$22$ & & $512$ & $512$ & $6$ & $8$ & $9$ & $2$ & $51.14$ & $42.69$ & $32.64$ & $21.04$ & $77.93$ & $65.93$ & $54.06$ & $36.52$  & \REVISION{$278.00$} \\%278002194
			$23$ & & $512$ & $512$ & $7$ & $8$ & $9$ & $2$ & $49.86$ & $42.24$ & $32.97$ & $21.47$ & $\mathbf{80.03}$ & $67.26$ & $54.79$ & $35.02$  & \REVISION{$322.92$} \\%322919445
% 			$13$ & & $128$ & $512$ & $5$ & $8$ & $9$ & $2$ & $49.86$ & $42.46$ & $31.97$ & $22.02$ & $75.88$ & $64.28$ &  $52.94$ & $34.97$ & $19124751$ \\
% 			$13$ & & $128$ & $512$ & $5$ & $8$ & $9$ & $2$ & $$ & $$ & $$  & $$ & $$ &  $$ & $$ & $$ & $15164943$ \\
% 			$13$ & & $100$ & $512$ & $5$ & $8$ & $9$ & $2$ & $$ & $$ & $$ & $$ & $$ & $$ &  $$ & $$ & $13467519$ \\
			$24$ & & $512$ & $256$ & $5$ & $8$ & $9$ & $2$ & $50.69$ & $43.96$ & $35.34$ & $\mathbf{24.34}$ & $77.61$ & $66.36$ & $54.81$ & $37.07$ & \REVISION{$233.08$} \\%233084943
			$25$ & & $512$ & $128$ & $5$ & $8$ & $9$ & $2$ & $48.56$ & $41.64$ & $33.34$ & $22.19$ & $76.36$ & $65.83$ & $53.79$ & $35.62$ & \REVISION{$233.08$} \\%233084943
			$26$ & & $128$ & $512$ & $3$ & $8$ & $9$ & $2$ & $44.64$ & $38.24$ & $31.54$ & $22.32$ & $69.81$ & $58.96$ & $49.54$ & $33.87$ & $9.52$ \\%9516297
			$27$ & & $162$ & $512$ & $3$ & $8$ & $9$ & $2$ & $44.21$ & $38.72$ & $31.22$ & $22.44$ & $72.26$ & $62.88$ & $51.81$ & $35.49$ & $14.99$ \\%14990841
			\hline
			$28$ & CBP & $-$ & $-$ & $1$  & $32$ & $-$ & $-$ & $-$ & $\mathit{27.31}$ & $\mathit{24.79}$ & $\mathit{19.10}$ & $-$ & $\mathit{43.64}$ & $\mathit{37.40}$ & $\mathit{25.59}$  & $15.34$ \\%15335458
            % $29$ & MS-2D-TAN (Shared) & $512$ & $512$ & $5$  & $8$ & $9$ & $2$ & $52.64$ & $44.34$ & $35.92$ & $24.64$ & $78.36$ &  $67.83$ &  $56.86$ & $38.39$ & $62.88$ \\
		    \hline
		\end{tabular}
	\end{center}
	\label{tab:hyperparameters}
\end{table*}

\subsubsection{Number of Scales}
\label{sec:ablation_scales}
In this section, we evaluate the performance under different number of scales $K$.
We vary the number of scales $K$ from $1$ to $7$ in our MS-2D-TAN model. The results are shown in Table~\ref{tab:hyperparameters} (row $17-23$). We observe that, increasing $K$ from $1$ to $5$ brings improvements ($25.49$ $v.s.$ $30.42$ $v.s.$ $34.09$ $v.s.$ $35.04$ $v.s.$ $35.77$ in $Rank5@0.5$). This observation is also consistent with the theoretical upper bound, as listed in Table~\ref{tab:hyperparameters} (Row $3-7$). Here, the upper bound represents the performance of an ideal model that can provide a correct prediction on all the sampled video clips. The upper bound is smaller than $100\%$ since the sampling of video moments may not cover all possible results. 
We also observe that further increasing the scale $K$ from $5$ to $7$ does not benefit the overall performance. Since the upper bounds (Row $7-9$) are already saturated, further increasing moment candidates may introduce additional data imbalance problem.
% In general, we find that $S=5$ achieve the best performance under $K=9$ and $L=2$.

\subsubsection{Number of Anchors}

In this section, we evaluate the performance under different number of anchors $A$. As mentioned in Sec.~\ref{sec:ablation_scales}, the upper bounds are vital to the overall performance. However, it is not the only factor.
With the same upper bounds (row $2$, $7$, $10$), we also vary the anchor size and the number of scales, as shown in Table~\ref{tab:ablation_study} (row $11$, $12$, $14$ and $21$).
From the experiments, we find that using $K=5$ and $A=8$ (row $21$) achieves the best performance in the TACoS evaluation metrics.
Using more maps with fewer anchors for each map may not provide sufficient context information at each scale (row $12$ $v.s.$ row $14$). 
Meanwhile, using less maps with more anchors at each map involves less long range context at each scale due to the smaller receptive field. (row $11$ $v.s.$ row $21$).

\subsubsection{Kernel and Layer Setting}

In this section, we evaluate different kernel size $\kappa$ and number of layers $L$.
With the same number of scales ($S=5$), we vary the kernel size and layers at each scale, as shown in Table~\ref{tab:hyperparameters} (row $13-16$ and row $21$).
There are several observations.
First, we compare the model with $\REVISION{\kappa}=1,L=1$ (row $13$) and $\REVISION{\kappa}=9,L=1$  (row $14$). Row $13$ predicts scores for each moment independently and row $14$ jointly considers other neighboring moments in the 2D temporal map. The significant improvement from these two rows verify the importance of modeling context information from adjacent moments. 
Further enlarging the receptive field can achieve better performance (row $15$, $16$, and $21$).
Comparing row $15$, $16$ and $21$ with the same receptive field size, we observe that the performance benefits more from larger kernel with fewer layers (row $16$ and row $21$) compared to small kernel with more layers (row $15$).

\subsubsection{Effectiveness of Window Size}

In this section, we compare the performance with different sliding window sizes ($N$) during training in Table~\ref{tab:hyperparameters} (row $24$, $25$ and $21$). We observe that larger window size can benefit the overall performance, since more context moments are observed during training.
By using $512$ clips extracted by C3D features (row $21$), we can cover approximate $96.42\%$ video length in average. And therefore it achieves the best performance compared with the row $25$ (covering $87.57\%$) and row $24$ (covering $67.47\%$).

\subsubsection{Effectiveness of Hidden State Size}

In this section, we evaluate the performance under different hidden state size $H$, as shown in (Row $26$, $27$, and $21$).
We observe that with larger number of hidden size, our model can achieve better performance. 
Even with about $\frac{2}{3}$ parameters of the previous best approach CBP~\cite{wang2020temporally}, our model can achieve superior performance (Row $27$ $v.s.$ $28$).

\subsubsection{Number of Candidate Moments}

Proposal based methods can benefit from a larger number of moments.
In this section, we verify whether the superior performance of our proposed MS-2D-TAN comes from the larger number of moments.
The number of moment candidates is a vital factor in moment localization models.
We first tune this factor in our MS-2D-TAN approach, and show its impacts on final performance. Next, we compare the previous best approach CBP~\cite{wang2020temporally} with respect to this factor.
The upper bound performance of proposal based methods is decided by the selection of moment candidates. 
Therefore, we evaluate our MS-2D-TAN with the same moment selection of previous state-of-the-art proposal based methods.
For a fair comparison, we compare CBP with our MS-2D-TAN on TACoS .
In CBP (Row $27$), each clip corresponds to $32$ anchors, while in the MS-2D-TAN with $A=8$ and $\REVISION{K}=3$ (Row $26$), the corresponding anchors are a subset of the CBP anchors ($16$ anchors).
With approximately the same number of parameters and half less number of anchors, our model outperforms the previous best approach by a large margin. 

\begin{figure*}[t!]
	\begin{subfigure}[t]{0.5\textwidth}
		\centering
		\includegraphics[width=\linewidth]{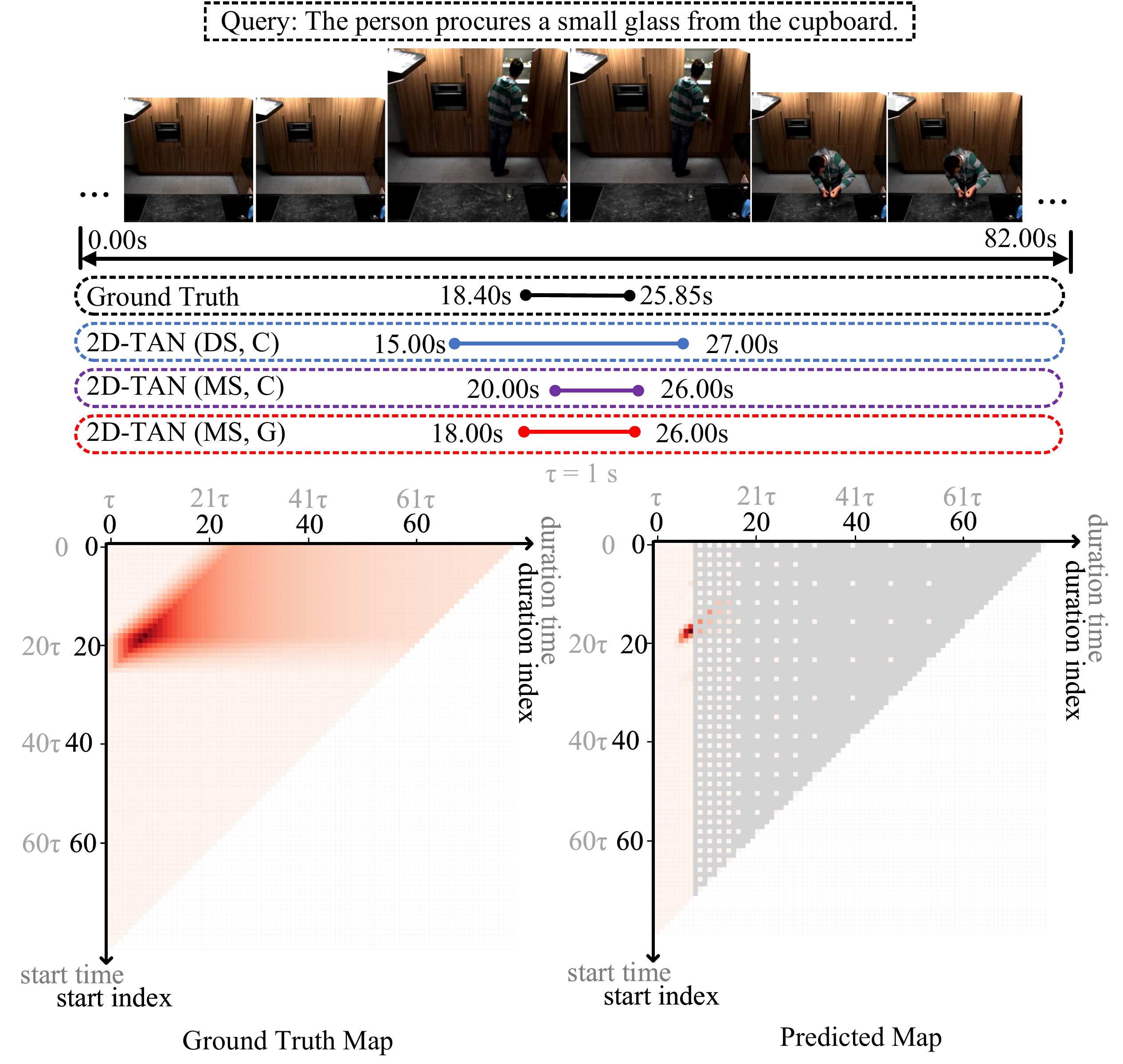}
		\caption{}\label{fig:example_a}		
	\end{subfigure}
% 	\quad
	\begin{subfigure}[t]{0.5\textwidth}
		\centering
		\includegraphics[width=\linewidth]{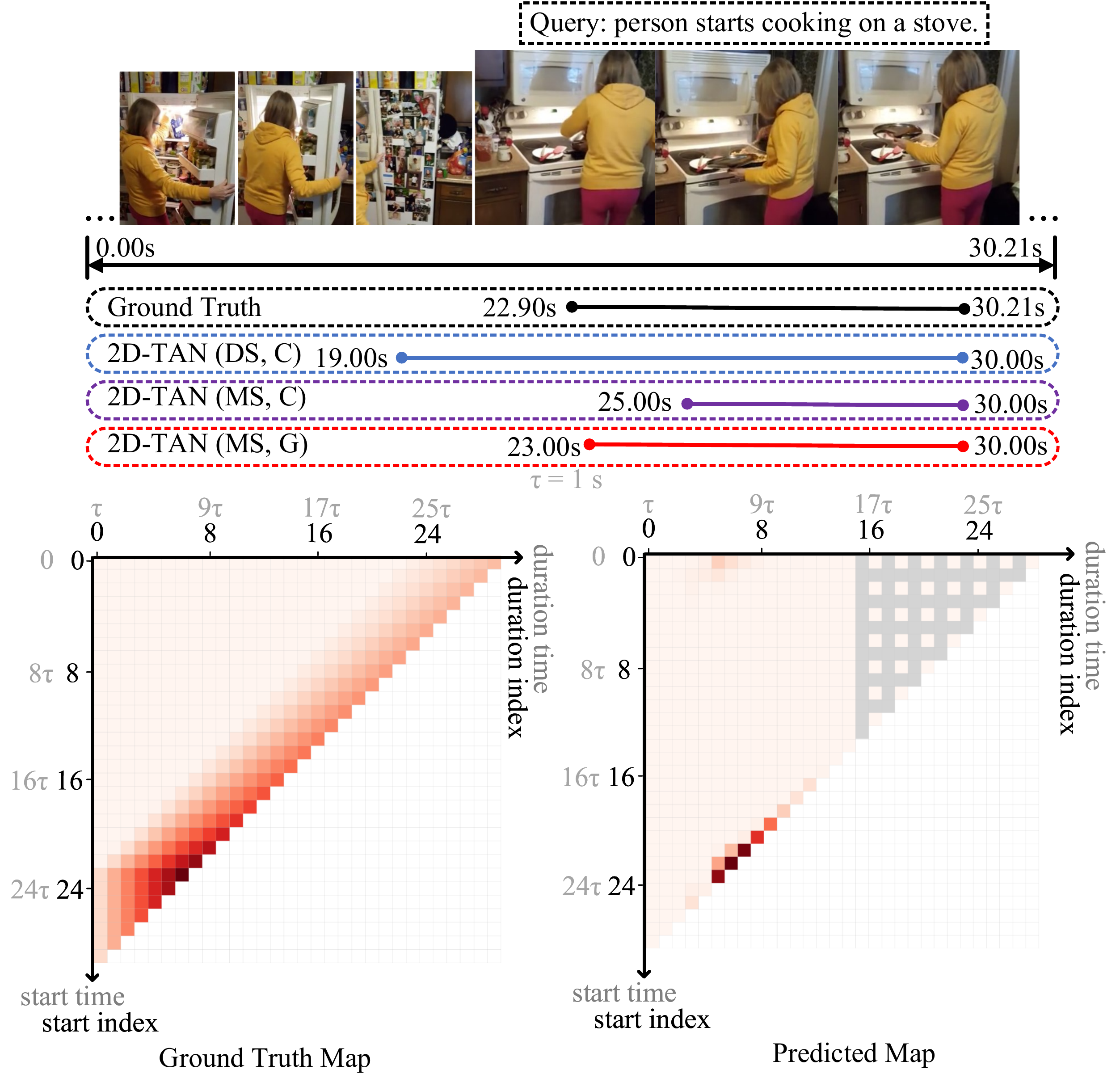}
		\caption{}\label{fig:example_b}
	\end{subfigure}
	\begin{subfigure}[t]{0.5\textwidth}
		\centering
		\includegraphics[width=\linewidth]{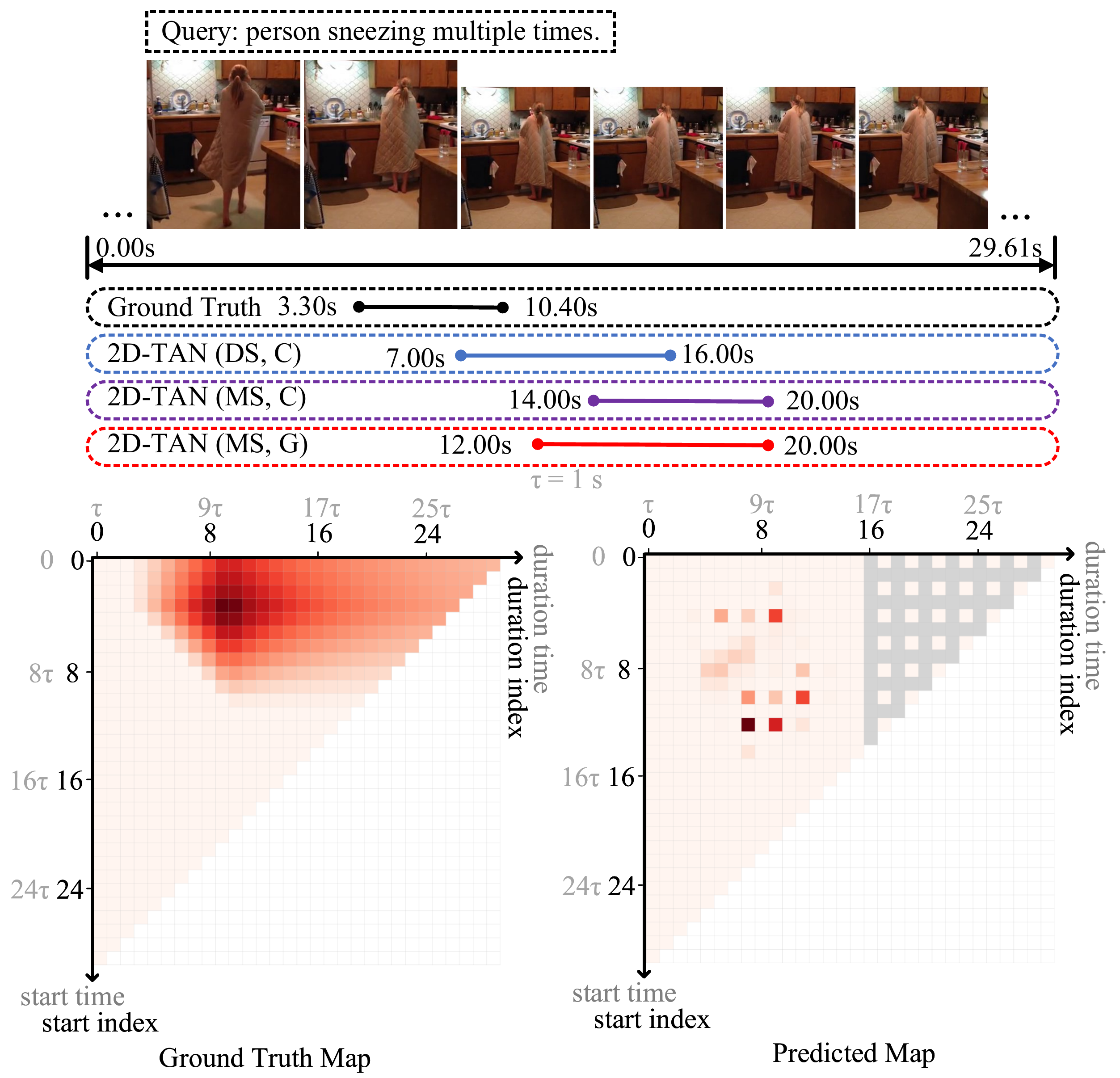}
		\caption{}\label{fig:example_c}
	\end{subfigure}
	\begin{subfigure}[t]{0.5\textwidth}
		\centering
		\includegraphics[width=\linewidth]{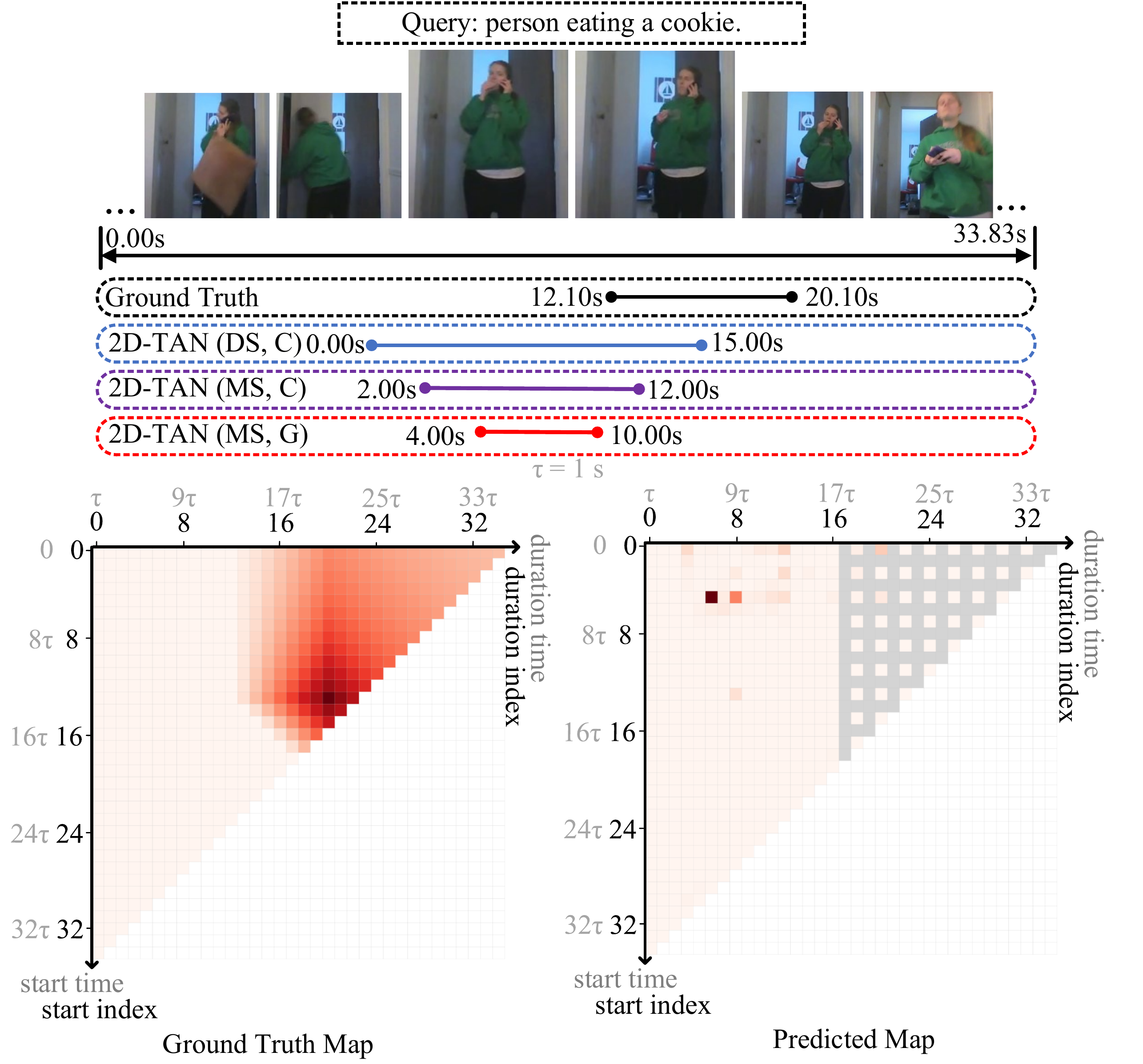}
		\caption{}\label{fig:example_d}
	\end{subfigure}
	\caption{\REVISION{Prediction examples of our model and the baselines. The map is predicted by our full model.}} %The predicted map is the result of our full model.}}
	\label{fig:examples}
\end{figure*}

\subsection{Qualitative Analysis}

We provide some qualitative examples to validate the effectiveness of our MS-2D-TAN. 
\REVISION{We first demonstrate two success cases in Fig.~\ref{fig:examples} (a - b).} The sparse multi-scale map exploits richer context information with a larger receptive field compared to the dense single-scale map. 
By combining with gated convolution, the location of the desired moment can be more precise.
We also visualize the predicted scores of our model. 
Comparing with the ground truth map, we can observe that the predicted map can correctly localize the target moment, while the surrounding moments also have high predicted scores. This demonstrates that our model can learn discriminative features from the neighboring moments.

\REVISION{
We further present two failure cases. In Fig.~\ref{fig:examples} (c), all models failed, since it's hard to localize ``sneezing'' without hearing the sound. In Fig.~\ref{fig:examples} (d), all models failed to localize ``eating a cookie''. 
% We believe further exploiting the audio modality and the object detector will benefit MS-2D-TAN. 
\SECONDRREVISION{Due to the lack of audio modality and object category information, MS-2D-TAN is not able to correctly localize these queries shown in Figure $6$. In the future, we would like to explore the multi-modal information and study how to interact multi-modal 2D temporal maps with sentence queries.}
% Besides, the sparse multi-scale modeling can reduce the computational cost, especially for long videos (Fig~\ref{fig:examples}a).
}

\section{Conclusion}
In this paper, we study the problem of moment localization with natural language and present a novel
Multi-Scale 2D Temporal \REVISION{Adjacency} Networks (MS-2D-TAN) method.
The core idea is to retrieve a moment on the multi-scale two-dimensional temporal map, which considers adjacent moment candidates as the temporal context. MS-2D-TAN is capable of encoding adjacent temporal context, while learning discriminative feature for matching video moments with the sentence query.
Our model is simple in design and achieves competitive performance in comparison with other state-of-the-art methods on three benchmark datasets.

\section{Acknowledgement}
This work is supported in part by NSF awards IIS-1704337, IIS-1722847, and IIS-1813709, as well as our corporate sponsors.

\bibliography{reference}

\begin{thebibliography}{10}
\providecommand{\url}[1]{#1}
\csname url@samestyle\endcsname
\providecommand{\newblock}{\relax}
\providecommand{\bibinfo}[2]{#2}
\providecommand{\BIBentrySTDinterwordspacing}{\spaceskip=0pt\relax}
\providecommand{\BIBentryALTinterwordstretchfactor}{4}
\providecommand{\BIBentryALTinterwordspacing}{\spaceskip=\fontdimen2\font plus
\BIBentryALTinterwordstretchfactor\fontdimen3\font minus
  \fontdimen4\font\relax}
\providecommand{\BIBforeignlanguage}[2]{{%
\expandafter\ifx\csname l@#1\endcsname\relax
\typeout{** WARNING: IEEEtran.bst: No hyphenation pattern has been}%
\typeout{** loaded for the language `#1'. Using the pattern for}%
\typeout{** the default language instead.}%
\else
\language=\csname l@#1\endcsname
\fi
#2}}
\providecommand{\BIBdecl}{\relax}
\BIBdecl

\bibitem{SSN2017ICCV}
Y.~Zhao, Y.~Xiong, L.~Wang, Z.~Wu, X.~Tang, and D.~Lin, ``Temporal action
  detection with structured segment networks,'' in \emph{ICCV}, 2017.

\bibitem{chen2019relation}
P.~Chen, C.~Gan, G.~Shen, W.~Huang, R.~Zeng, and M.~Tan, ``Relation attention
  for temporal action localization,'' \emph{TMM}, 2019.

\bibitem{zeng2019breaking}
R.~Zeng, C.~Gan, P.~Chen, W.~Huang, Q.~Wu, and M.~Tan, ``Breaking
  winner-takes-all: Iterative-winners-out networks for weakly supervised
  temporal action localization,'' \emph{TIP}, 2019.

\bibitem{xu2020g}
M.~Xu, C.~Zhao, D.~S. Rojas, A.~Thabet, and B.~Ghanem, ``G-tad: Sub-graph
  localization for temporal action detection,'' in \emph{CVPR}, 2020.

\bibitem{lin2021learning}
C.~Lin, C.~Xu, D.~Luo, Y.~Wang, Y.~Tai, C.~Wang, J.~Li, F.~Huang, and Y.~Fu,
  ``Learning salient boundary feature for anchor-free temporal action
  localization,'' in \emph{CVPR}, 2021.

\bibitem{hasan2016learning}
M.~Hasan, J.~Choi, J.~Neumann, A.~K. Roy-Chowdhury, and L.~S. Davis, ``Learning
  temporal regularity in video sequences,'' in \emph{CVPR}, 2016.

\bibitem{song2015tvsum}
Y.~Song, J.~Vallmitjana, A.~Stent, and A.~Jaimes, ``{TVSum}: Summarizing web
  videos using titles,'' in \emph{CVPR}, 2015.

\bibitem{chu2015video}
W.-S. Chu, Y.~Song, and A.~Jaimes, ``Video co-summarization: Video
  summarization by visual co-occurrence,'' in \emph{CVPR}, 2015.

\bibitem{gao2017tall}
J.~Gao, C.~Sun, Z.~Yang, and R.~Nevatia, ``{TALL}: Temporal activity
  localization via language query,'' in \emph{ICCV}, 2017.

\bibitem{hendricks17iccv}
L.~A. Hendricks, O.~Wang, E.~Shechtman, J.~Sivic, T.~Darrell, and B.~Russell,
  ``Localizing moments in video with natural language,'' in \emph{ICCV}, 2017.

\bibitem{escorcia2019temporal}
V.~Escorcia, M.~Soldan, J.~Sivic, B.~Ghanem, and B.~Russell, ``Temporal
  localization of moments in video collections with natural language,''
  \emph{arXiv preprint arXiv:1907.12763}, 2019.

\bibitem{zhao2021cascaded}
Y.~Zhao, Z.~Zhao, Z.~Zhang, and Z.~Lin, ``Cascaded prediction network via
  segment tree for temporal video grounding,'' in \emph{CVPR}, 2021.

\bibitem{lin2020moment}
Z.~Lin, Z.~Zhao, Z.~Zhang, Z.~Zhang, and D.~Cai, ``Moment retrieval via
  cross-modal interaction networks with query reconstruction,'' \emph{TIP},
  2020.

\bibitem{lei2018tvqa}
J.~Lei, L.~Yu, M.~Bansal, and T.~L. Berg, ``{TVQA}: Localized, compositional
  video question answering,'' in \emph{EMNLP}, 2018.

\bibitem{Shao_2018_ECCV}
D.~Shao, Y.~Xiong, Y.~Zhao, Q.~Huang, Y.~Qiao, and D.~Lin, ``Find and focus:
  Retrieve and localize video events with natural language queries,'' in
  \emph{ECCV}, 2018.

\bibitem{duan2018weakly}
X.~Duan, W.~Huang, C.~Gan, J.~Wang, W.~Zhu, and J.~Huang, ``Weakly supervised
  dense event captioning in videos,'' in \emph{NeurIPS}, 2018.

\bibitem{gella2018dataset}
S.~Gella, M.~Lewis, and M.~Rohrbach, ``A dataset for telling the stories of
  social media videos,'' in \emph{EMNLP}, 2018.

\bibitem{Ge_2019_WACV}
R.~Ge, J.~Gao, K.~Chen, and R.~Nevatia, ``{MAC}: Mining activity concepts for
  language-based temporal localization,'' in \emph{WACV}, 2019.

\bibitem{liu2018attentive}
M.~Liu, X.~Wang, L.~Nie, X.~He, B.~Chen, and T.-S. Chua, ``Attentive moment
  retrieval in videos,'' in \emph{SIGIR}, 2018.

\bibitem{song2018val}
X.~Song and Y.~Han, ``{VAL}: Visual-attention action localizer,'' in
  \emph{PCM}, 2018.

\bibitem{2DTAN_2020_AAAI}
S.~Zhang, H.~Peng, J.~Fu, and J.~Luo, ``Learning 2d temporal adjacent networks
  for moment localization with natural language,'' in \emph{AAAI}, 2020.

\bibitem{lin2017single}
T.~Lin, X.~Zhao, and Z.~Shou, ``Single shot temporal action detection,'' in
  \emph{ACMMM}, 2017.

\bibitem{chen2018temporally}
J.~Chen, X.~Chen, L.~Ma, Z.~Jie, and T.-S. Chua, ``Temporally grounding natural
  sentence in video,'' in \emph{EMNLP}, 2018.

\bibitem{zhang2019cross}
Z.~Zhang, Z.~Lin, Z.~Zhao, and Z.~Xiao, ``Cross-modal interaction networks for
  query-based moment retrieval in videos,'' in \emph{SIGIR}, 2019.

\bibitem{yuan2019semantic}
Y.~Yuan, L.~Ma, J.~Wang, W.~Liu, and W.~Zhu, ``Semantic conditioned dynamic
  modulation for temporal sentence grounding in videos,'' in \emph{NeurIPS},
  2019.

\bibitem{zhang2020span}
H.~Zhang, A.~Sun, W.~Jing, and J.~T. Zhou, ``Span-based localizing network for
  natural language video localization,'' in \emph{ACL}, 2020.

\bibitem{yuan2019to}
Y.~Yuan, T.~Mei, and W.~Zhu, ``To find where you talk: Temporal sentence
  localization in video with attention based location regression,'' in
  \emph{AAAI}, 2019.

\bibitem{ghosh2019excl}
S.~Ghosh, A.~Agarwal, Z.~Parekh, and A.~Hauptmann, ``{ExCL}: Extractive clip
  localization using natural language descriptions,'' in \emph{NAACL}, 2019.

\bibitem{zeng2020dense}
R.~Zeng, H.~Xu, W.~Huang, P.~Chen, M.~Tan, and C.~Gan, ``Dense regression
  network for video grounding,'' \emph{CVPR}, 2020.

\bibitem{liu2018crossmodal}
M.~Liu, X.~Wang, L.~Nie, Q.~Tian, B.~Chen, and T.-S. Chua, ``Cross-modal moment
  localization in videos,'' in \emph{ACMMM}, 2018.

\bibitem{mun2020local}
J.~Mun, M.~Cho, and B.~Han, ``Local-global video-text interactions for temporal
  grounding,'' in \emph{CVPR}, 2020.

\bibitem{Rodriguez_2020_WACV}
C.~Rodriguez, E.~Marrese-Taylor, F.~S. Saleh, H.~LI, and S.~Gould,
  ``Proposal-free temporal moment localization of a natural-language query in
  video using guided attention,'' in \emph{WACV}, March 2020.

\bibitem{chen2020rethinking}
L.~Chen, C.~Lu, S.~Tang, J.~Xiao, D.~Zhang, C.~Tan, and X.~Li, ``Rethinking the
  bottom-up framework for query-based video localization,'' in \emph{AAAI},
  2020.

\bibitem{lu2019debug}
C.~Lu, L.~Chen, C.~Tan, X.~Li, and J.~Xiao, ``Debug: A dense bottom-up
  grounding approach for natural language video localization,'' in
  \emph{EMNLP}, 2019.

\bibitem{jiang2019cross}
B.~Jiang, X.~Huang, C.~Yang, and J.~Yuan, ``Cross-modal video moment retrieval
  with spatial and language-temporal attention,'' in \emph{ICMR}, 2019.

\bibitem{he2019read}
D.~He, X.~Zhao, J.~Huang, F.~Li, X.~Liu, and S.~Wen, ``Read, watch, and move:
  Reinforcement learning for temporally grounding natural language descriptions
  in videos,'' in \emph{AAAI}, 2019.

\bibitem{wang2019language}
W.~Wang, Y.~Huang, and L.~Wang, ``Language-driven temporal activity
  localization: A semantic matching reinforcement learning model,'' in
  \emph{CVPR}, 2019.

\bibitem{Hahn2019tripping}
M.~Hahn, A.~Kadav, J.~M. Rehg, and H.~P. Graf, ``Tripping through time:
  Efficient localization of activities in videos,'' in \emph{CVPR Workshop},
  2019.

\bibitem{wu2020tree}
J.~Wu, G.~Li, S.~Liu, and L.~Lin, ``Tree-structured policy based progressive
  reinforcement learning for temporally language grounding in video,'' in
  \emph{AAAI}, 2020.

\bibitem{xu2019multilevel}
H.~Xu, K.~He, B.~A. Plummer, L.~Sigal, S.~Sclaroff, and K.~Saenko, ``Multilevel
  language and vision integration for text-to-clip retrieval,'' in \emph{AAAI},
  2019.

\bibitem{chen2019semantic}
S.~Chen and Y.-G. Jiang, ``Semantic proposal for activity localization in
  videos via sentence query,'' in \emph{AAAI}, 2019.

\bibitem{zhang2019man}
D.~Zhang, X.~Dai, X.~Wang, Y.-F. Wang, and L.~S. Davis, ``{MAN}: Moment
  alignment network for natural language moment retrieval via iterative graph
  adjustment,'' in \emph{CVPR}, 2019.

\bibitem{dehmamy2019understanding}
N.~Dehmamy, A.-L. Barab{\'a}si, and R.~Yu, ``Understanding the representation
  power of graph neural networks in learning graph topology,'' in
  \emph{NeurIPS}, 2019.

\bibitem{liu2018temporal}
B.~Liu, S.~Yeung, E.~Chou, D.-A. Huang, L.~Fei-Fei, and J.~Carlos~Niebles,
  ``Temporal modular networks for retrieving complex compositional activities
  in videos,'' in \emph{ECCV}, 2018.

\bibitem{wang2020temporally}
J.~Wang, L.~Ma, and W.~Jiang, ``Temporally grounding language queries in videos
  by contextual boundary-aware prediction,'' in \emph{AAAI}, 2020.

\bibitem{Beltagy2020Longformer}
I.~Beltagy, M.~E. Peters, and A.~Cohan, ``Longformer: The long-document
  transformer,'' \emph{arXiv:2004.05150}, 2020.

\bibitem{sukhbaatar2019adaptive}
S.~Sukhbaatar, E.~Grave, P.~Bojanowski, and A.~Joulin, ``Adaptive attention
  span in transformers,'' \emph{ACL}, 2019.

\bibitem{lin2019bmn}
T.~Lin, X.~Liu, X.~Li, E.~Ding, and S.~Wen, ``{BMN}: Boundary-matching network
  for temporal action proposal generation,'' in \emph{ICCV}, 2019.

\bibitem{lin2020fast}
C.~Lin, J.~Li, Y.~Wang, Y.~Tai, D.~Luo, Z.~Cui, C.~Wang, J.~Li, F.~Huang, and
  R.~Ji, ``Fast learning of temporal action proposal via dense boundary
  generator.'' in \emph{AAAI}, 2020.

\bibitem{pennington2014glove}
J.~Pennington, R.~Socher, and C.~D. Manning, ``Glove: Global vectors for word
  representation,'' in \emph{EMNLP}, 2014.

\bibitem{hochreiter1997long}
S.~Hochreiter and J.~Schmidhuber, ``Long short-term memory,'' \emph{Neural
  computation}, vol.~9, no.~8, pp. 1735--1780, 1997.

\bibitem{yu2019free}
J.~Yu, Z.~Lin, J.~Yang, X.~Shen, X.~Lu, and T.~S. Huang, ``Free-form image
  inpainting with gated convolution,'' in \emph{ICCV}, 2019.

\bibitem{krishna2017dense}
R.~Krishna, K.~Hata, F.~Ren, L.~Fei-Fei, and J.~C. Niebles, ``Dense-captioning
  events in videos,'' in \emph{ICCV}, 2017.

\bibitem{regneri2013grounding}
M.~Regneri, M.~Rohrbach, D.~Wetzel, S.~Thater, B.~Schiele, and M.~Pinkal,
  ``Grounding action descriptions in videos,'' \emph{TACL}, 2013.

\bibitem{rohrbach2012script}
M.~Rohrbach, M.~Regneri, M.~Andriluka, S.~Amin, M.~Pinkal, and B.~Schiele,
  ``Script data for attribute-based recognition of composite activities,'' in
  \emph{ECCV}, 2012.

\bibitem{simonyan2014very}
K.~Simonyan and A.~Zisserman, ``Very deep convolutional networks for
  large-scale image recognition,'' in \emph{ICPR}, 2015.

\bibitem{tran2015learning}
D.~Tran, L.~Bourdev, R.~Fergus, L.~Torresani, and M.~Paluri, ``Learning
  spatiotemporal features with 3d convolutional networks,'' in \emph{ICCV},
  2015.

\bibitem{carreira2017quo}
J.~Carreira and A.~Zisserman, ``Quo vadis, action recognition? a new model and
  the kinetics dataset,'' in \emph{CVPR}, 2017.

\bibitem{deng2009imagenet}
J.~Deng, W.~Dong, R.~Socher, L.-J. Li, K.~Li, and L.~Fei-Fei, ``Imagenet: A
  large-scale hierarchical image database,'' in \emph{CVPR}, 2009.

\bibitem{karpathy2014large}
A.~Karpathy, G.~Toderici, S.~Shetty, T.~Leung, R.~Sukthankar, and L.~Fei-Fei,
  ``Large-scale video classification with convolutional neural networks,'' in
  \emph{CVPR}, 2014.

\bibitem{sigurdsson2016hollywood}
G.~A. Sigurdsson, G.~Varol, X.~Wang, A.~Farhadi, I.~Laptev, and A.~Gupta,
  ``Hollywood in homes: Crowdsourcing data collection for activity
  understanding,'' in \emph{ECCV}, 2016.

\bibitem{kingma2014adam}
D.~P. Kingma and J.~Ba, ``Adam: A method for stochastic optimization,'' in
  \emph{ICPR}, 2015.

\bibitem{wu2018multi}
A.~Wu and Y.~Han, ``Multi-modal circulant fusion for video-to-language and
  backward,'' in \emph{IJCAI}, 2018.

\bibitem{ren2015faster}
S.~Ren, K.~He, R.~Girshick, and J.~Sun, ``Faster r-cnn: Towards real-time
  object detection with region proposal networks,'' in \emph{NeurIPS}, 2015.

\bibitem{hendricks18emnlp}
L.~A. Hendricks, O.~Wang, E.~Shechtman, J.~Sivic, T.~Darrell, and B.~Russell,
  ``Localizing moments in video with temporal language.'' in \emph{EMNLP},
  2018.

\bibitem{zhang2019exploiting}
S.~Zhang, J.~Su, and J.~Luo, ``Exploiting temporal relationships in video
  moment localization with natural language,'' in \emph{ACMMM}, 2019.

\end{thebibliography}
\bibliographystyle{IEEEtran}

\begin{IEEEbiography}
    [{\includegraphics[width=1in,height=1.25in,keepaspectratio]{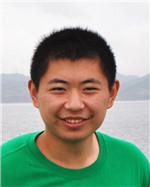}}]{Songyang Zhang}
    received the BS degree from
Southeast University, China, in 2015 and MS degree from Zhejiang University, China, in 2018. He is currently working toward the PhD degree at University of Rochester.
His research interest includes moment localization with natural language, unsupervised grammar induction, skeleton-based action recognition, etc.
\end{IEEEbiography}

\begin{IEEEbiography}
    [{\includegraphics[width=1in,height=1.25in,clip,keepaspectratio]{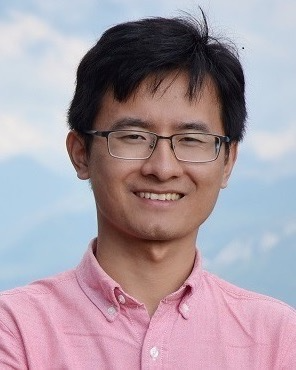}}]{Houwen Peng}
    is a researcher working on computer vision and deep learning at Microsoft Research as of 2018. Before that he was a senior engineer at Qualcomm AI Research. He received Ph.D. from NLPR, Instituation of Automation, Chinese Academy of Sciences in 2016. From 2015 to 2016, he worked as a visiting research scholar at Temple University. His research interest includes neural architecture search and design, video object tracking, segmentation and detection, video moment localization, saliency detection, etc.
\end{IEEEbiography}

\begin{IEEEbiography}
    [{\includegraphics[width=1in,height=1.25in,clip,keepaspectratio]{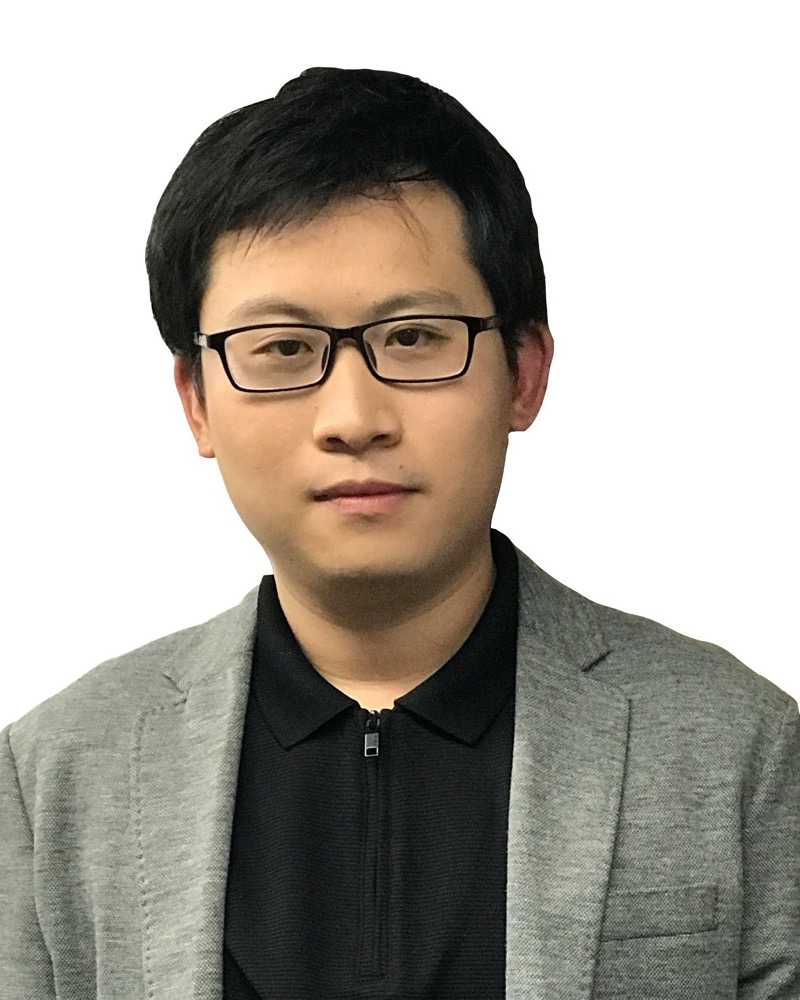}}]{Jianlong Fu}
    is currently a Senior Research Manager with the Multimedia Search and Mining Group, Microsoft Research Asia (MSRA). He is now leading Multimedia Search and Mining Group, focusing on computer vision, image graphics, vision and language, especially on fine-grained image/video recognition and detection, multimedia content editing, personal media experience of browsing, searching, sharing and storytelling. He has shipped core technologies to a number of Microsoft products, including Windows, Office, Bing Multimedia Search, Azure Media Service, XiaoIce, etc.
\end{IEEEbiography}

\begin{IEEEbiography}
    [{\includegraphics[width=1in,height=1.25in,keepaspectratio]{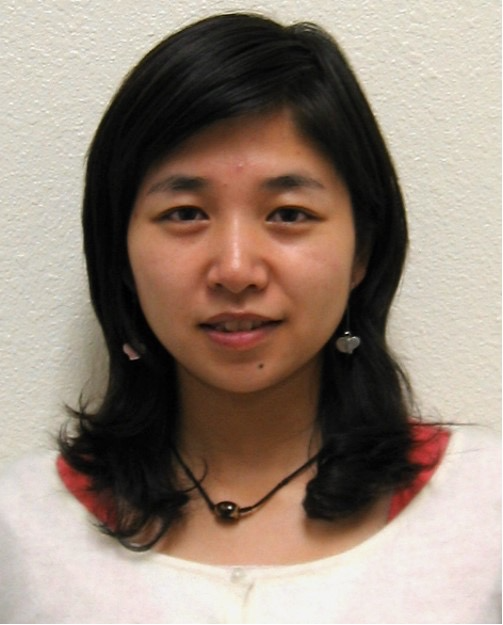}}]{Yijuan(Lucy) Lu} is a Principal Scientist at Microsoft Azure AI, where she worked on OCR, object detection and video understanding in the recent two years. Prior to joining Microsoft, she was an associate professor in the Department of Computer Science at Texas State University. Her major publications appear in leading publication venues in multimedia and computer vision research.  She was the First Place Winner in many challenging retrieval competitions in Eurographics for many years. She received 2015 Texas State Presidential Distinction Award and 2014 College Achievement Award. She also received the Best Paper award from ICME 2013 and ICIMCS 2012. She has obtained many competitive external grants from NSF, US Army, US Department of Defense and Texas Department of Transportation.
\end{IEEEbiography}

\begin{IEEEbiography}
    [{\includegraphics[width=1in,height=1.25in,keepaspectratio]{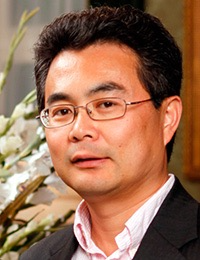}}]{Jiebo Luo} (S93, M96, SM99, F09) is a Professor of Computer Science at the University of Rochester which he joined in 2011 after a prolific career of fifteen years at Kodak Research Laboratories. He has authored nearly 500 technical papers and holds over 90 U.S. patents. His research interests include computer vision, NLP, machine learning, data mining, computational social science, and digital health. He has been involved in numerous technical conferences, including serving as a program co-chair of ACM Multimedia 2010, IEEE CVPR 2012, ACM ICMR 2016, and IEEE ICIP 2017, as well as a general co-chair of ACM Multimedia 2018. He has served on the editorial boards of the IEEE Transactions on Pattern Analysis and Machine Intelligence (TPAMI), IEEE Transactions on Multimedia (TMM), IEEE Transactions on Circuits and Systems for Video Technology (TCSVT), IEEE Transactions on Big Data (TBD), ACM Transactions on Intelligent Systems and Technology (TIST), Pattern Recognition, Knowledge and Information Systems (KAIS), Machine Vision and Applications, and Journal of Electronic Imaging. He is the current Editor-in-Chief of the IEEE Transactions on Multimedia. Professor Luo is also a Fellow of ACM, AAAI, SPIE, and IAPR.
\end{IEEEbiography}

\end{document}